\documentclass[twoside]{article}

\usepackage[accepted]{aistats2020}
\usepackage[round]{natbib}

\usepackage{subcaption}
\usepackage{algorithm}
\usepackage{algorithmic}
\usepackage{amssymb}  
\usepackage{amsfonts}       % blackboard math symbols
\usepackage{amsmath}
\usepackage{amsthm}
\usepackage{mathtools}
\usepackage{bm}
\usepackage{bbm}
\usepackage{changepage}
\usepackage[makeroom]{cancel}
\usepackage{sidecap}
\usepackage{longtable}
\usepackage{pifont}
\usepackage[dvipsnames]{xcolor}

\usepackage{textcomp,    % for \textlangle and \textrangle macros
	xspace}
\usepackage{pgfplots}
\usepackage{tikz} 
\usepackage{bigints}

\pgfplotsset{compat=newest}
\usetikzlibrary{decorations.pathmorphing} % noisy shapes
\usetikzlibrary{fit} % fitting shapes to coordinates
\usetikzlibrary{backgrounds} % drawing the background after the foreground
\usetikzlibrary{pgfplots.groupplots}
\usetikzlibrary{positioning}

\newcommand{\myalg}{NOPG}

\newcommand*\de{\mathop{}\!\mathrm{d}}

\DeclareMathOperator*{\EV}{\mathbb{E}}

\newcommand{\gradt}[0]{\nabla_{\theta}}

\newcommand{\Aset}[0]{\mathcal{A}}
\newcommand{\Sset}[0]{\mathcal{S}}
\newcommand{\svec}[0]{\mathbf{s}}
\newcommand{\xvec}[0]{\mathbf{x}}
\newcommand{\avec}[0]{\mathbf{a}}
\newcommand{\rvec}[0]{\mathbf{r}}
\newcommand{\qvec}[0]{\mathbf{q}_{\pi}}

\newcommand{\phivec}[0]{\bm{\phi}}
\newcommand{\pis}[0]{\pi_{\theta}}

\newcommand{\approxx}[1]{\hat{#1}}
\newcommand{\grad}[1]{\nabla_{#1}}
\newcommand{\kerres}{\bm{\Lambda}_{\pi}}
\newcommand{\pargrad}[1]{\frac{\partial}{\partial {#1}}}

\newcommand{\state}[0]{\mathbf{s}}
\newcommand{\action}[0]{\mathbf{a}}
\newcommand{\nextstate}[0]{\mathbf{s}'}
\newcommand{\Trans}[3]{p({#1}|{#2},{#3})}

\newcommand{\defeq}[0]{\coloneqq}

\newcommand{\bvec}{\mathbf{b}}
\newcommand{\zvec}{\mathbf{z}}
\newcommand{\Avec}{\mathbf{A}}

\newcommand{\lvec}{\mathbf{l}}
\newcommand{\hvec}{\mathbf{h}}

\newcommand{\cmark}{\textcolor{OliveGreen}{\ding{51}}}%
\newcommand{\xmark}{\textcolor{BrickRed}{\ding{55}}}%
\newcommand{\tlow}{\textcolor{OliveGreen}{low}}
\newcommand{\thigh}{\textcolor{BrickRed}{high}}

\DeclareMathOperator\erf{erf}
\newcommand{\e}[1]{e^{\textstyle #1}}

\newtheorem{definition}{Definition}
\newtheorem{theorem}{Theorem}

\newtheorem{proposition}{Proposition}

\usepackage{hyperref}

\begin{document}



\twocolumn[

%\aistatstitle{A Sample Efficient Nonparametric Off-Policy Policy Gradient}
\aistatstitle{A Nonparametric Off-Policy Policy Gradient}
% use \And
\aistatsauthor{Samuele Tosatto\textsuperscript{1}
\And
Jo\~{a}o Carvalho\textsuperscript{1}
\And
Hany Abdulsamad\textsuperscript{1}
\And 
Jan Peters\textsuperscript{1,2}}

\aistatsaddress{\textsuperscript{1}Technische Universit\"at Darmstadt \\ \textsuperscript{2}Max Planck Institute for Intelligent Systems\\
\{tosatto, carvalho, abdulsamad, peters\}@ias.tu-darmstadt.de}
]

\newlength\figureheight
\newlength\figurewidth
		
	\begin{abstract}
		% 1> What is our problem?
		Reinforcement learning (RL) algorithms still suffer from high sample complexity despite outstanding recent successes. The need for intensive interactions with the environment is especially observed in many widely popular policy gradient algorithms that perform updates using on-policy samples.
		% 2> Why is this important?
		The price of such inefficiency becomes evident in real world scenarios such as interaction-driven robot learning, where the success of RL has been rather limited.
		% 3> Our suggestion to solve this!
		We address this issue by building on the general sample efficiency of off-policy algorithms. With nonparametric regression and density estimation methods we construct a \textsl{nonparametric Bellman equation} in a principled manner, which allows us to obtain closed-form estimates of the value function, and to analytically express the \textsl{full} policy gradient.
		% 4> Why does our contribution help
		We provide a theoretical analysis of our estimate to show that it is consistent under mild smoothness assumptions and empirically show that our approach has better sample efficiency than state-of-the-art policy gradient methods.
	\end{abstract}
	
	\section{Introduction}
	\label{introduction}
%	The domain of reinforcement learning has experienced an overwhelming progress in recent years, achieving outstanding results on hard high-dimensional tasks \citep{mnih2015human,silver2017mastering,ecoffet2019go}. However, in comparison to advances in simulated environments and computer games, progress in transferring recent insights to the field of reinforcement learning for robotics and robot learning has been significantly conservative \citep{kober2013reinforcement,gu2017deep}, where poor sample efficiency combined with the extraordinary costs of system interactions in a real-world application makes such transfer virtually intractable. Our objective is to address this issue in a theoretically principled way in the context of off-policy policy-gradient learning techniques.
	
	%\setlength\figureheight{6.6cm}
	\setlength\figureheight{5.25 cm}
	\setlength\figurewidth{0.64 \columnwidth}
	\begin{figure}
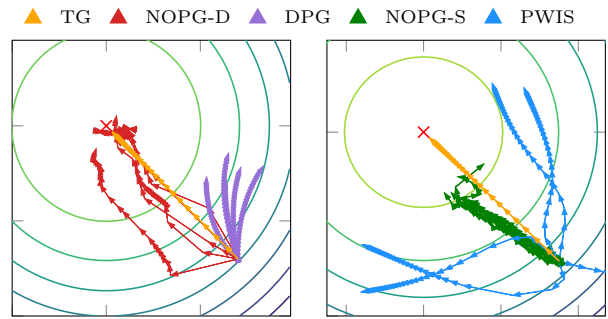

		\centering	\begin{tikzpicture}

\definecolor{color8}{rgb}{0.59,0.44,0.84}  % DPG
\definecolor{color9}{rgb}{1,0.647058823529412,0}  % True
\definecolor{color10}{rgb}{0.83921568627451,0.152941176470588,0.156862745098039}  % NOPG-D
\definecolor{color6}{rgb}{0.117647058823529,0.564705882352941,1}  % PWIS

\begin{axis}[
height=2cm,
width=5cm,
hide axis,
xmin=10,
xmax=50,
ymin=0,
ymax=0.5,
legend columns=5,
legend entries={{TG}, {NOPG-D}, {DPG}, {NOPG-S}, {PWIS}},
legend style={at={(1.0,0.1)}, anchor=north, draw=none, mark options={scale=1.}, font=\scriptsize, column sep=1.5ex, line width=2 pt},
]

\addlegendimage{only marks, mark=triangle, color9}
\addlegendimage{only marks, mark=triangle, color10}
\addlegendimage{only marks, mark=triangle, color8}
\addlegendimage{only marks, mark=triangle, green!50.19607843137255!black}
\addlegendimage{only marks, mark=triangle, color6}

\end{axis}

\end{tikzpicture}
		\input{plots/LQR/deterministic.tikz}
		\input{plots/LQR/stochastic.tikz}
		\caption{Example showing the bias of offline-DPG (left) and the variance of PWIS-G(PO)MDP (right) in the policy-parameter space of a 2d-LQR setting. Both algorithms diverge while they move away from the ``on-policy'' region. Our method in its deterministic and stochastic versions, NOPG-D and NOPG-S, shows better approximations of the true gradient (TG).}
		\label{figure:gradient}
	\end{figure}
	\setlength\figureheight{5.25 cm}
	\setlength\figurewidth{0.65 \columnwidth}
	
	Reinforcement learning has made overwhelming progress in recent years \citep{mnih_human-level_2015,haarnoja_soft_2018,schulman_trust_2015}. However, the vast majority of reinforcement learning approaches are on-policy algorithms with limited applicability to real world scenarios, due to high sample complexity. In contrast, off-policy techniques are theoretically more sample efficient, because they decouple the procedures of data acquisition and policy update, allowing for the possibility of sample-reuse and safe interaction with the environment. These two properties are of high importance when developing algorithms for real robots.  However, classical off-policy algorithms like Q-learning with function approximation and fitted Q-iteration \citep{ernst_tree-based_2005,riedmiller_neural_2005} are not guaranteed to converge \citep{baird_residual_1995,lu_non-delusional_2018}, and allow only discrete actions. More recent semi-gradient\footnote{We adopt the terminology from \cite{imani_off-policy_2018}.} off-policy techniques, like Off-PAC \citep{degris_off-policy_2012-1} and DDPG \citep{silver_deterministic_2014,lillicrap_continuous_2016} often perform sub-optimally, especially when the collected data is strongly off-policy, due to the biased semi-gradient update \citep{fujimoto_off-policy_2019}. Off-policy algorithms based on importance sampling \citep{shelton_policy_2001,meuleau_exploration_2001,peshkin_learning_2002} deliver an unbiased estimate of the gradient but suffer from high variance and are generally only applicable with stochastic policies. Moreover, they require the full knowledge of the behavioral policy, making them unsuitable when data stems from a human demonstrator. Additionally, model-based approaches like PILCO \citep{deisenroth_pilco:_2011} may be considered to be off-policy. Such probabilistic nonlinear trajectory optimizers are limited to the finite-horizon domain and suffer from coarse approximations when propagating the state distribution through time. To address all previously highlighted issues in state-of-the-art off-policy approaches, we propose a new algorithm, the nonparametric off-policy policy gradient (NOPG), a full-gradient estimate based on the closed-form solution of a nonparametric Bellman equation. Furthermore, we avoid the high variance of importance sampling techniques and allow for the use of human demonstrations. Figure~\ref{figure:gradient} qualitatively compares the gradient estimate of NOPG compared to that of semi-parametric approaches (DPG) and path-wise importance sampling (PWIS) techniques. Furthermore, unlike other nonparametric approaches like PILCO, our approach allows for multimodal state-transitions, and can handle the infinite-horizon setting. For empirical validation, we evaluate our approach on a number of classical control tasks. The results highlight the sample efficiency of our approach.

	\section{Notation and Background}
	Consider the reinforcement learning problem of an agent interacting with a given environment, as abstracted by a Markov decision process (MDP) and defined over the tuple $(\Sset, \Aset, \gamma, P, R)$ where $\Sset \equiv \mathbb{R}^{d_s}$ is the state space, $\Aset \equiv \mathbb{R}^{d_a}$ the action space and $\gamma \in [0,1)$ the discount factor. The transition probability from a state $\state$ to $\nextstate$ given an action $\action$ is governed by the conditional density $\Trans{\nextstate}{\state}{\action}$. The stochastic reward signal $R$ for a state-action pair $(\state, \action) \in \Sset \times \Aset$ is drawn from a distribution $R(\state, \action)$ with mean value $r(\state, \action)$. The policy $\pi$, parameterized by $\theta$, is a stochastic or deterministic mapping from $\Sset$ onto $\Aset$. Our objective is to maximize the expected return
	\begin{equation}
		J_{\pi}= \EV\left[\sum_{t=0}^{\infty} \gamma^{t} R_t\right] \label{eq:return} .
	\end{equation}
	 Following \cite{sutton_policy_2000}, we define $\mu_{\pi}(\svec)\!=\!\sum_{t=0}^\infty \gamma^t p(\state_t | \state_0, \pi)$ as the state-visitation function induced by the policy $\pis$. A state-action value function $Q_{\pi}(\state, \action)$ maps the state-action pair onto $\mathbb{R}$ and represents the expected discounted cumulative return following the policy $\pis$. The state value function $V_{\pi}$ is the expectation of $Q_{\pi}$ under $\pis$.
	
	\textbf{Policy Gradient Theorem.} Objective \eqref{eq:return} can be maximized via gradient ascent. The gradient of $J_{\pi}$ w.r.t. the policy parameters $\theta$ is

	\begin{equation*}
	\nabla_{\theta}J_{\pi} \!=\! \int\displaylimits_{\Sset}\int\displaylimits_{\Aset}\! \mu_{\pi}(\svec)\pi_{\theta}(\avec|\svec)Q_{\pi}(\svec,\avec)\nabla_{\theta}\log \pi_{\theta}(\avec | \svec) \de \avec \de \svec,
	\end{equation*}
	as stated in the policy gradient theorem \citep{sutton_policy_2000}. In a direct episodic on-policy setting, the expected return $Q_{\pi}$ can be estimated under the current state-action visitation $\mu_{\pi}(\svec)\pi_{\theta}(\avec|\svec)$ via Monte-Carlo episodic rollouts of the current policy \citep{williams_simple_1992}. This technique, however, may require excessive interactions with the environment, since the return and the expectations need to be approximated after each policy update.
	
	\textbf{Off-Policy Semi-Gradient.} The off-policy policy gradient theorem was the first proposed off-policy actor-critic algorithm \citep{degris_off-policy_2012-1}. Since then it has inspired a series of state-of-the-art off-policy algorithms \citep{silver_deterministic_2014,lillicrap_continuous_2016}. Nonetheless, its important to note that this theorem and its successors, introduce two approximations to the original policy gradient theorem. Firstly, semi-gradient approaches consider a modified discounted infinite-horizon return objective $\approxx{J}_{\pi} = \int\rho_{\beta}(\svec)V_{\pi}(\svec) \de \state$, where $\rho_{\beta}(\svec)$ is the stationary state distribution under the behavioral policy $\pi_{\beta}$. Secondly, the gradient estimate is modified to be
	\begin{align}
	\gradt \approxx{J}_{\pi} & = \gradt\int_{\Sset} \rho_{\beta}(\state) V_{\pi}(\state)\de \state \nonumber \\
	& = \gradt \int_{\Sset} \rho_{\beta}(\state) \int_{\Aset} \pi_{\theta}(\action|\state) Q_{\pi}(\state, \action)\de \action \de \state  \nonumber                                                 \\
	& = \int\displaylimits_{\Sset} \rho_{\beta}(\state) \int\displaylimits_{\Aset} \underbrace{\gradt \pi_{\theta}(\action|\state) Q_{\pi}(\state, \action)}_{\text{A}} \nonumber \\
	& \quad \quad + \underbrace{\pi_{\theta}(\action|\state) \gradt Q_{\pi}(\state, \action)}_{\text{B}} \de \action \de \state \label{equation:qgrad} \\
	& \approx \int\displaylimits_{\Sset} \rho_{\beta}(\state) \int\displaylimits_{\Aset} \gradt \pi_{\theta}(\action|\state) Q_{\pi}(\state, \action) \de \action\de \state, \nonumber
	\end{align}
	where the term $\text{B}$ related to the derivative of $Q_{\pi}$ is ignored. In all fairness, the authors provide a proof that this biased gradient, or \textsl{semi-gradient}, still converges to the optimal policy in a discrete MDP setting \citep{degris_off-policy_2012-1,imani_off-policy_2018}. 

	\textbf{Path-Wise Importance Sampling (PWIS).} One way to obtain an unbiased estimate of the policy gradient in an off-policy scenario is to re-weight every trajectory via importance sampling \citep{meuleau_exploration_2001,shelton_policy_2001, peshkin_learning_2002}. An example of the gradient estimation via G(PO)MDP with importance sampling is given by
	\begin{align}
		\nabla_{\theta} J_{\pi} = \mathbb{E}\left[\sum_{t=0}^{T-1} \rho_t Q_{\pi}(\state_t, \action_t) \nabla_{\theta} \log \pi_{\theta}(\action_t | \state_t) \right] \label{eq:pwis},
	\end{align}
	where $\rho_t = \prod_{z=0}^{t} \pi_{\theta}(\action_z | \state_z)/\pi_{\beta}(\action_z | \state_z)$.
	This technique applies only to stochastic policies and requires the knowledge of the behavioral policy $\pi_\beta$. Moreover, Equation~\eqref{eq:pwis} shows that PWIS needs a trajectory-based dataset, since it needs to keep track of the past in the correction term $\rho_z$, hence introducing more restrictions on its applicability.
	Additionally, importance sampling suffers from high variance \citep{owen_monte_2013}, which grows multiplicatively in the number of steps (Equation~\ref{eq:pwis}). Despite these difficulties, many interesting recent advances have helped to make PWIS more reliable. For example, \cite{imani_off-policy_2018} propose a trade-off between PWIS and semi-gradient approaches, \cite{metelli_policy_2018} argue for the use of a surrogate objective which accounts for the variance of the estimate and \cite{liu_breaking_2018, liu_off-policy_2019} apply importance sampling to the state distribution instead of the trajectories.
	
	\section{Nonparametric Off-Policy Policy Gradient}
	We introduce a new offline off-policy approach with a full-gradient estimate that does not suffer from the drawbacks of importance sampling and semi-gradient algorithms. Starting from a nonparametric Bellman equation, we derive an analytical expression of the gradient for deterministic and stochastic policies. 
	Nonparametric Bellman equations have been developed in a number of prior works. \cite{ormoneit_kernel-based_2002,xu_kernel-based_2007,engel_reinforcement_2005} used nonparametric models such as Gaussian processes for approximate dynamic programming. \cite{taylor_kernelized_2009} have shown that these methods differ mainly in their use of regularization. \cite{kroemer_non-parametric_2011} provided a Bellman equation using kernel density-estimation and a general overview over nonparametric dynamic programming. In contrast to prior work, our formulation preserves the dependency on the policy enabling the computation of the policy gradient in closed-form. Moreover, we upper-bound the bias of the Nadaraya-Watson kernel regression to prove that our value function estimate is consistent w.r.t. the classical Bellman equation under smoothness assumptions. \label{nppg}
	We focus on the maximization of the average return in the infinite horizon case.
	\begin{definition}
		The discounted infinite-horizon objective is defined by $J_{\pi} = \int \mu_{0}(\state)V_{\pi}(\state)\de \state$, where $ \mu_{0}$ is the initial state distribution. Under a stochastic policy the objective is subject to the constraint
		\label{definition:objective}
		\begin{align}
		V_{\pi}(\state)&\!=\!\int_{\Aset} \pi_{\theta}(\action|\state) \bigg(r(\state,\action) \nonumber
		\\ & \quad \quad +\gamma\!\int_{\Sset} V_{\pi}(\nextstate)p(\nextstate|\state, \action)\de \nextstate \bigg)\!\de \action,
		\end{align}
		while in the case of a deterministic policy the constraint is given as
		\begin{equation*}
		V_{\pi}(\state) = r(\state,\pi_{\theta}(\state)) + \gamma \int_{\Sset} V_{\pi}(\nextstate)p(\nextstate|\state, \pi_{\theta}(\state))\de \nextstate.
		\end{equation*}
	\end{definition}
	Maximizing the objective in Definition~\ref{definition:objective} analytically is not possible, excluding special cases such as under  linear-quadratic assumptions \citep{borrelli_predictive_2017}. Extracting an expression for the gradient of $J_{\pi}$ w.r.t. the policy parameters $\theta$ is also not straightforward given the infinite set of possibly non-convex constraints represented in the recursion over $V_{\pi}$. Nevertheless, it is possible to transform the constraints in Definition~\ref{definition:objective} to a finite set of linear constraints via nonparametric modeling, thus leading to an expression of the value function with simple algebraic manipulation \citep{kroemer_non-parametric_2011}.
	
	\paragraph{Nonparametric Modeling.} Assume a set of $n$ observations $D\!\equiv\!\{\state_i, \action_i, r_i, {\nextstate}_i\}_{i=1}^{n}$ sampled from interaction with an environment, with $\state_i, \action_i \sim \beta(\cdot, \cdot)$, $\state_i' \sim p(\cdot |\state_i, \action_i)$ and $r_i \sim R(\state_i, \action_i)$ . We define the kernels $\psi:\Sset\times\Sset \to \mathbb{R}^+$, $\varphi:\Aset\times\Aset\to \mathbb{R}^+$ and $\phi:\Sset\times\Sset\to\mathbb{R}^+$, as normalized, symmetric and positive definite functions with bandwidths $\hvec_{\varphi}, \hvec_{\phi}, \hvec_{\psi}$ respectively. We define $\psi_i(\state) = \psi(\state, \state_i)$, $\varphi_i(\action) = \varphi(\action, \action_i)$, and $\phi_i(\state) = \phi(\state, {\nextstate}_i)$. Following \cite{kroemer_non-parametric_2011}, the mean reward $r(\state, \action)$ and the transition conditional $p(\nextstate|\state, \action)$ are approximated by the Nadaraya-Watson regression \citep{nadaraya_estimating_1964,watson_smooth_1964} and kernel density estimation, respectively
	\begin{align}
	\approxx{r}(\state, \action) &\! \defeq \! \frac{\sum_{i=1}^n\psi_i(\state)\varphi_i(\action)r_i}{ \sum_{i=1}^n\psi_i(\state)\varphi_i(\action)}\nonumber \\ 
	p(\nextstate| \state, \action) &\! \approx\!  \frac{\approxx{p}(\state', \action, \state)}{\approxx{p}(\action, \state)} \defeq  \approxx{p}(\nextstate | \state, \action) \nonumber %\label{eq:transkernel},
	\end{align}
	where $\approxx{p}(\nextstate, \state, \action) = 1/n \sum_i \phi_i(\nextstate)\psi_i(\state)\varphi_i(\action)$ and $\approxx{p}(\state, \action) = 1/n \sum_i \psi_i(\state)\varphi_i(\action)$.
	
	Inserting the reward and transition kernels into the Bellman Equation for the case of stochastic policy, we obtain the nonparametric Bellman equation (NPBE)
	\begin{align}
	\approxx{V}_{\pi}(\svec) &\!=\!\int_{\Aset} \! \pis(\avec|\svec)\bigg(\approxx{r}(\state, \action)\!+\!\gamma\!\int_{\Sset} \approxx{V}_{\pi}(\svec')\approxx{p}(\svec'|\svec, \avec) \de \svec'\bigg)\!\de \avec \nonumber \\
	& \!=\!\sum_i \int_{\Aset}  \frac{\pis(\avec|\svec) \psi_i(\state)\varphi_i(\action)}{\sum_j\psi_j(\state)\varphi_j(\action)}\de \avec \nonumber \\
	& \quad \quad \quad \times \bigg(r_i + \gamma \int_{\Sset} \phi_i(\svec')\approxx{V}_{\pi}(\svec') \de \svec' \bigg). \label{equation:npbe} 
	\end{align}
	Equation~\eqref{equation:npbe} can be conveniently expressed in matrix form by introducing the vector of responsibilities $\varepsilon_i(\state)\!=\!\int \pis(\avec|\svec)\! \psi_i(\state)\varphi_i(\action)/\sum_j\psi_j(\state)\varphi_j(\action)\de \avec$, which assigns each state $\state$ a weight relative to its distance to a sample $i$ under the current policy.
	\begin{definition}{}
		\label{definition:npbe}
		The nonparametric Bellman equation on the dataset $D$ is formally defined as
		\begin{equation}
		\approxx{V}_{\pi}(\svec)\!=\!\bm{\varepsilon}_{\pi}^{\intercal}(\svec) \left(\rvec + \gamma \int_{\Sset} \phivec(\svec')\approxx{V}_{\pi}(\svec') \de \svec' \right), \label{eq:formalnpbe}
		\end{equation}
		\begin{align}
			& \text{with }  \phivec(\svec)  \!=\![\phi_1(\svec),\dots, \phi_n(\svec)]^{\intercal}, \rvec  \!=\![r_1,\dots, r_n]^{\intercal}, \nonumber \\
			& \bm{\varepsilon}_{\pi}(\svec)   \!=\![\varepsilon_1^\pi(\svec),\! \dots,\!\varepsilon_n^\pi(\svec)]^{\intercal}, \nonumber \\
			& \varepsilon^{\pi}_i(\svec)\!  = \! \begin{cases}
			\int \pis(\avec|\svec) \frac{\psi_i(\state)\varphi_i(\action)}{\sum_j\psi_j(\state)\varphi_j(\action)}\de \avec & \text{if $\pi$ is stochastic} \\  \frac{\psi_i(\state)\varphi_i(\pis(\svec))}{\sum_j\psi_j(\state)\varphi_j(\pis(\svec))} & \text{otherwise.} \nonumber
			\end{cases} 
		\end{align}
		% 	\begin{equation}
		% 		\begin{cases}
		% 			\varepsilon^{\pi}_i(\svec) \defeq \int \pis(\avec|\svec) \frac{\psi_i(\state)\varphi_i(\action)}{\sum_j\psi_j(\state)\varphi_j(\action)}\de \avec & \text{if $\pi$ is stochastic}, \\
		% 			\varepsilon^{\pi}_i(\svec)  \defeq \frac{\psi_i(\state)\varphi_i(\pis(\svec))}{\sum_j\psi_j(\state)\varphi_j(\pis(\svec))}                                & \text{otherwise.}
		% 		\end{cases}\nonumber
		% 	\end{equation}
	\end{definition}
	
	From Equation~\eqref{eq:formalnpbe} we deduce that the value function must be of the form $\bm{\varepsilon}_{\pi}^\intercal(\state)\qvec$, indicating that it can also be seen as a form of Nadaraya-Watson kernel regression,
	\begin{equation}
	\bm{\varepsilon}_{\pi}^\intercal(\state)\qvec = \bm{\varepsilon}_{\pi}^{\intercal}(\svec) \left(\rvec + \gamma \int_{\Sset} \phivec(\svec')\bm{\varepsilon}_{\pi}^{\intercal}(\svec)\qvec \de \svec' \right). \label{eq:npbewresp}
	\end{equation}
	Notice that every $\qvec$ which satisfies 
	\begin{equation}
	\qvec = \rvec + \gamma \int_{\Sset} \phivec(\svec')\bm{\varepsilon}_{\pi}^{\intercal}(\svec)\qvec \de \svec' \label{eq:npbeworesp}
	\end{equation}
	also satisfies Equation~\eqref{eq:npbewresp}. Theorem~\ref{theorem:fixedpoint} demonstrates that the algebraic solution of Equation~\eqref{eq:npbeworesp} is the \textsl{only} solution of the nonparametric Bellman Equation~\eqref{eq:formalnpbe}. 
	
	\begin{theorem}{}
		\label{theorem:fixedpoint}
		The nonparametric Bellman equation has a unique fixed-point solution 
		\begin{equation}
			\approxx{V}^*_{\pi}(\svec) \defeq \bm{\varepsilon}^{\intercal}_{\pi}(\svec)\bm{\Lambda}_\pi^{-1}\rvec, \nonumber
		\end{equation} with $\kerres \defeq I-\gamma\approxx{\mathbf{P}}_{\pi}$ and $\approxx{\mathbf{P}}^{\pi}_{i,j} \defeq \int\phi_i(\svec')\varepsilon^{\pi}_j(\svec')\de \svec'$, where $\bm{\Lambda}_{\pi}$ is always invertible since $\approxx{\mathbf{P}}_{\pi}$ is a stochastic matrix and $0 \leq \gamma < 1$.
	\end{theorem}
	Proof of Theorem~\ref{theorem:fixedpoint} is provided in the supplementary material.
	
	\paragraph{Policy Gradient.} With the closed-form solution of $\approxx{V}^*_{\pi}(\svec)$ from Theorem~\ref{theorem:fixedpoint} it is possible to compute the analytical gradient of $J_{\pi}$ w.r.t. the policy parameters
	\begin{align}
	\grad{\theta} \approxx{V}_{\pi}^*(\svec) &=  \bigg( \pargrad{\theta}\bm{\varepsilon}_{\pi}^{\intercal}(\svec)\bigg)\kerres^{-1} \mathbf{r}  + \bm{\varepsilon}_{\pi}^{\intercal}(\svec) \pargrad{\theta} \kerres^{-1} \mathbf{r}  \nonumber \\
	%&= & \bigg( \pargrad{\theta}\bm{\varepsilon}_{\pi}^{\intercal}(\svec)\bigg)\kerres^{-1} \mathbf{r}  - \bm{\varepsilon}_{\pi}^{\intercal}(\svec)\kerres^{-1}\bigg(\pargrad{\theta}\kerres\bigg)\kerres^{-1}\mathbf{r} \nonumber \\
	&=  \underbrace{\bigg( \pargrad{\theta}\bm{\varepsilon}_{\pi}^{\intercal}(\svec)\bigg)\kerres^{-1} \mathbf{r}}_{\text{A}} \nonumber \\ &  \quad \quad + \underbrace{ \gamma \bm{\varepsilon}_{\pi}^{\intercal}(\svec)\kerres^{-1}\bigg(\pargrad{\theta} \approxx{\mathbf{P}}_{\pi}\bigg)\kerres^{-1} \mathbf{r}}_{\text{B}}. \label{equation:kergradv}
	\end{align}
	Substituting the result of Equation~\eqref{equation:kergradv} into the return specified in Definition~\ref{definition:objective}, introducing $\bm{\varepsilon}_{\pi,0}^{\intercal} \defeq \int\mu_0(\svec)\bm{\varepsilon}_{\pi}^{\intercal}(\svec) \de \svec$, $\mathbf{q}_{\pi} = \bm{\Lambda}^{-1}_\pi \rvec$, and $\bm{\mu}_{\pi} = \bm{\Lambda}_{\pi}^{-\intercal} \bm{\varepsilon}_{\pi,0}$ we obtain
	\begin{equation}
	\grad{\theta} \approxx{J}_{\pi} = \bigg( \pargrad{\theta}\bm{\varepsilon}_{\pi,0}^{\intercal}\bigg)\mathbf{q}_{\pi}  + \gamma \bm{\mu}_{\pi}^{\intercal}\bigg(\pargrad{\theta} \approxx{\mathbf{P}}_{\pi}\bigg)\mathbf{q}_{\pi}, \label{equation:algorithmicgradient}
	\end{equation}
	where $\mathbf{q}_{\pi}$ and $\bm{\mu}_{\pi}$ can be estimated via conjugate gradient to avoid the inversion of $\bm{\Lambda}_\pi$.
	
	The terms $\text{A}$ and $\text{B}$ in Equation~\eqref{equation:kergradv} correspond to the terms in Equation~\eqref{equation:qgrad}. In contrast to semi-gradient actor-critic methods, where the gradient bias is affected by both the critic bias and the semi-gradient approximation \citep{imani_off-policy_2018,fujimoto_off-policy_2019}, our estimate is the \textsl{full gradient} and the only source of bias is introduced by the estimation of $\approxx{V}_{\pi}$, which we analyze in Section~\ref{sec:nonparaestimation}.
	The term $\bm{\mu}_\pi$ can be interpreted as the support of the state-distribution as it satisfies  $\bm{\mu}^{\intercal}_\pi =\bm{\varepsilon}_{\pi,0}^{\intercal} + \gamma\bm{\mu}_\pi^{\intercal} {\approxx{\mathbf{P}}_{\pi}}$. In Section~\ref{section:experiments}, more specifically in Figure~\ref{figure:muv}, we empirically show that $\varepsilon_{\pi}^{\intercal}(\svec)\bm{\mu}_\pi$ provides an estimate of the state distribution over the whole state-space. Implementation-wise, the quantities $\bm{\varepsilon}_{\pi,0}^{\intercal}$ and $\approxx{\mathbf{P}}^{\pi}_{i,j}$ are estimated via  Monte-Carlo sampling, which is unbiased but computationally demanding, or using other techniques such as unscented transform or numerical quadrature. The matrix $\approxx{\mathbf{P}}_{\pi}$ is of dimension $n \times n$, which can be memory-demanding. In practice, we notice that the matrix is often almost sparse. By taking advantage of conjugate gradient and sparsification we are able to achieve computational complexity of $\mathcal{O}(n^2)$ per policy update and memory complexity of $\mathcal{O}(n)$. A schematic of our implementation is summarized in Algorithm~\ref{alg:kbpgalg}.
	
	\begin{algorithm}[t]
		\caption{Nonparametric Off-Policy Policy Gradient}
		\label{alg:kbpgalg}
		\begin{algorithmic}
			\STATE \textbf{input:} dataset $\{\state_i, \action_i, r_i,\nextstate_i, \text{t}_i \}_{i=1}^{n}$ where $\text{t}_i$ indicates a terminal state, a policy $\pi_{\theta}$ and kernels $\psi, \phi, \varphi$ respectively for state, action and next-state.
			\WHILE{ \text{not converged}}
			\STATE Compute $\bm{\varepsilon}^{\intercal}_{\pi}(\svec)$ as in Definition~\ref{definition:npbe} and  $\bm{\varepsilon}_{\pi,0}^{\intercal} \defeq \int\mu_0(\svec)\bm{\varepsilon}_{\pi}^{\intercal}(\svec) \de \svec$.
			\STATE Estimate $\approxx{\mathbf{P}}_{\pi}$ as defined in Theorem~\ref{theorem:fixedpoint} using MC ($\phi(\svec)$ is a distribution).
			\STATE Set each row $i$ of $\approxx{\mathbf{P}}_{\pi}$ to $0$ if $\text{t}_i$ is a terminal state.
			\STATE Solve $\rvec = \bm{\Lambda}_\pi\mathbf{q}_{\pi}$ and $\bm{\varepsilon}_{\pi,0} = \bm{\Lambda}_{\pi}^{\intercal}\bm{\mu}_{\pi}$ for $\mathbf{q}_{\pi}$ and $\bm{\mu}_{\pi}$ using conjugate gradient.
			\STATE Update $\theta$ using Equation~\eqref{equation:algorithmicgradient}.
			\ENDWHILE
		\end{algorithmic}
	\end{algorithm}
	% \begin{definition}{}
	% 	The full off-policy policy gradient of Definition~\ref{definition:objective} is analytically written as
	% 	\begin{equation}
	% 		\grad{\theta}  \approxx{J}_{\pi} = \int\mu_0(\svec)\grad{\theta} \approxx{V}_{\pi}^*(\svec)\de \state = \bigg( \pargrad{\theta}\bm{\varepsilon}_{\pi,0}^{\intercal}\bigg)\kerres^{-1} \mathbf{r}  +  \gamma \bm{\varepsilon}_{\pi,0}^{\intercal}\kerres^{-1}\bigg(\pargrad{\theta} \approxx{\mathbf{P}}_{\pi}\bigg)\kerres^{-1} \mathbf{r} \label{equation:kergradj},
	% 	\end{equation}
	% 	where $\bm{\varepsilon}_{\pi,0}^{\intercal} \defeq \int\mu_0(\svec)\bm{\varepsilon}_{\pi}^{\intercal}(\svec) \de \svec$.
	% 	\label{definition:kergrad}
	% \end{definition}
	\section{Error Analysis of Nonparametric Estimates}
	\label{sec:nonparaestimation}
	Nonparametric estimates of the transition dynamics and reward enjoy favorable properties for an off-policy learning setting. A well-known asymptotic behavior of the Nadaraya-Watson kernel regression,
	\begin{align}
	& \EV \left[\lim_{n\to \infty}\approxx{f}_n(x) \right] - f(x)  \approx \nonumber\\
	& \quad\quad\quad  h^2_n \bigg(\frac{1}{2}f''(x) + \frac{f'(x){ \beta'(x)}}{ \beta(x)}\bigg) \int u^2 K(u)\de u, \nonumber %+o_p(h_n^2) \nonumber
	\end{align}
	shows how the bias is related to the regression function $f(x)$, as well as to the samples' distribution $\beta(x)$  \citep{fan_design-adaptive_1992,wasserman_all_2006}. 
	However, this asymptotic behavior is valid only for infinitesimal bandwidth, infinite samples ($h \to 0, nh \to \infty$) and requires the knowledge of the regression function and of the sampling distribution.
	
	In a recent work, we propose an upper bound of the bias that is also valid for finite bandwidths \citep{tosatto_upper_2020}. 
	We show that under some Lipschitz conditions, the bound of the Nadaraya-Watson kernel regression bias does not depend on the samples' distribution, which is a desirable property in off-policy scenarios. 
	The analysis is extended to multidimensional input space. For clarity of exposition, we report the main result in its simplest formulation, and later use it to infer the bound of the NPBE bias.
	%The estimation bias reported from  In Theorem~\ref{theorem:fixedpoint} we show how to derive a closed-form solution of the nonparametric Bellman equation. It is however important to understand the statistical properties of this result in order to quantify its usefulness. 
	%	In the following, we prove two theorems which show that the approximation of $\approxx{V}_{\pi}^*(\svec)$ is consistent w.r.t. to the true Bellman solution under reasonable assumptions. We start by analyzing the bias introduced by the Nadaraya-Watson kernel regression estimator given samples drawn from a distribution $\beta(\state)$. 
	%The bias introduced by $\beta$ is referred to as \textsl{design bias} \citep{wasserman2006all}. 
	%	In Theorem~\ref{theorem:biasnadaraya} we derive a novel design-bias-free upper-bound of the kernel regression estimator under some assumptions such as a ``smooth'' sampling density and Gaussian kernels. We consider this contribution important and argue that this property of the estimator is fundamental for true off-policy learning. 
	%To the best of our knowledge, this derivation is the first design-free upper-bound in the literature.
	\begin{theorem}{}
		\label{theorem:biasnadaraya}
		Let $f\!:\!\mathbb{R}^d\!\to\!\mathbb{R}$ be a Lipschitz continuous function with constant $L_{f}$.
		Assume a set $\{\xvec_i, y_i\}_{i=1}^n$ of i.i.d. samples from a log-Lipschitz distribution $\beta$ with a Lipschitz constant $L_{\beta}$. Assume $y_i = f(\xvec_i) + \epsilon_i$, where $f\!:\!\mathbb{R}^d\!\to\!\mathbb{R}$ and $\epsilon_i$ is i.i.d. and zero-mean. The bias of the Nadaraya-Watson kernel regression with Gaussian kernels in the limit of infinite samples $n\to \infty$ is bounded by
		\begin{equation*}
		\begin{split}
		& \left|\EV \left[\lim_{n\to \infty}\approxx{f}_n(\xvec) \right] - f(\xvec) \right| \leq \\
		& \quad \quad \quad \quad \frac{L_f \sum\limits_{k=1}^d \hvec_k \left(\prod\limits_{i\neq k}^d \chi_i\right)  \left( \frac{1}{\sqrt{2 \pi} } + \frac{L_{\beta}\hvec_k}{2} \chi_k \right) }{ \prod\limits_{i=1}^d \e{\frac{L_{\beta}^2 h_i^2}{2}}\left(1 - \erf\left(\frac{\hvec_i L_{\beta}}{\sqrt{2}} \right) \right)},\\
		\end{split}
		\end{equation*}
		where \begin{equation}
		\chi_i = e^{\frac{L_{\beta}^2\hvec^2_i}{2}}\bigg(1+\erf\bigg(\frac{\hvec_iL_\beta}{\sqrt{2}} \bigg)\bigg) \nonumber ,
		\end{equation}
		$\hvec > 0 \in \mathop{R}^d$ is the vector of bandwidths and $\erf{}$ is the error function.
	\end{theorem}
	
	%	Note that the estimate is unbiased in the limit of infinite samples and infinitesimal bandwidths $\{\hvec_i\}_{i=1}^n$. 
	%Moreover, the Lipschitz constant $L_{\beta}$ highlights the role the assumed smoothness of the sampling density plays. A full proof of this theorem can be found in the supplementary material.
	Building on Theorem~\ref{theorem:biasnadaraya} we show that the solution of the NPBE is consistent with the solution of the true Bellman equation. Moreover, although the bound is not affected directly by $\beta(\state)$, a smoother sample distribution $\beta(\state)$ plays favorably in the bias term (a low $L_{\beta}$ is preferred).
	\begin{theorem}{}
		\label{theorem:ultimate}
		Consider an arbitrary MDP $\mathcal{M}$ with a transition density $p$ and a stochastic reward function $R(\svec, \avec) = r(\svec,\avec) + \epsilon_{\svec, \avec}$, where $r(\svec,\avec)$ is a Lipschitz continuous function with $L_R$ constant and $\epsilon_{\svec, \avec}$ denotes zero-mean noise. Assume $|R(\svec, \avec)|\!\leq\!R_{\text{max}}$ and a dataset $D_n$ sampled from a log-Lipschitz distribution $\beta$ defined over the state-action space with Lipschitz constant $L_{\beta}$. Let $V_D$ be the unique solution of a nonparametric Bellman equation with Gaussian kernels $\psi,\varphi,\phi$ with positive bandwidths $\hvec_{\psi},\hvec_{\varphi}, \hvec_{\phi}$ defined over the dataset $\lim_{n\to\infty} D_n$. 
		%$V_D$ is an estimator of the fixed point $V^*$ of the classical Bellman equation defined over $\mathcal{M}$. 
		Assume $V_D$ to be Lipschitz continuous with constant $L_V$. The bias of such estimator is bounded by
		\begin{equation}
		\big|\overline{V}(\svec) - V^*(\svec)\big| \leq  \frac{1}{1-\gamma} \bigg(\text{A}_\text{Bias} + \gamma L_{V} \sum_{k=1}^{d_s}\frac{h_{\phi,k}}{\sqrt{2 \pi}} \bigg),
		\end{equation}
		where $\overline{V}(\svec) = \EV_{D}[V_D(\svec)]$, $V^*(\state)$ is the fixed point of the classic Bellman equation, $\text{A}_\text{Bias}$ is the bound of the bias provided in Theorem~\ref{theorem:biasnadaraya} with $L_f\!=\!L_R$, $\hvec\!=\![\hvec_{\psi},\hvec_{\varphi}]$ and $d\!=\!d_s\!+\!d_a$.\footnote{Complete proofs of the theorems and precise definitions can be found in the supplementary material.}
	\end{theorem}
	%\vspace{2em}
	Theorem~\ref{theorem:ultimate} shows that the value function provided by Theorem~\ref{theorem:fixedpoint} is consistent. Moreover, it is interesting to notice that the error can be decomposed in $A_\text{Bias}$, which is the bias component dependent on the reward's approximation, and the remaining term that depends on the smoothness of the value function and the bandwidth of $\phi$, which can be read as the error of the transition's model.
	
	The independence from the sampling distribution suggests that, under these assumptions, nonparametric estimation is particularly suited for off-policy setting, as the bias is not affected by different behavioral policies.  More in detail, the bound shows that smoother reward functions, state-transitions and sample distributions play favorably against the estimation bias.

	\begin{figure*}[t]
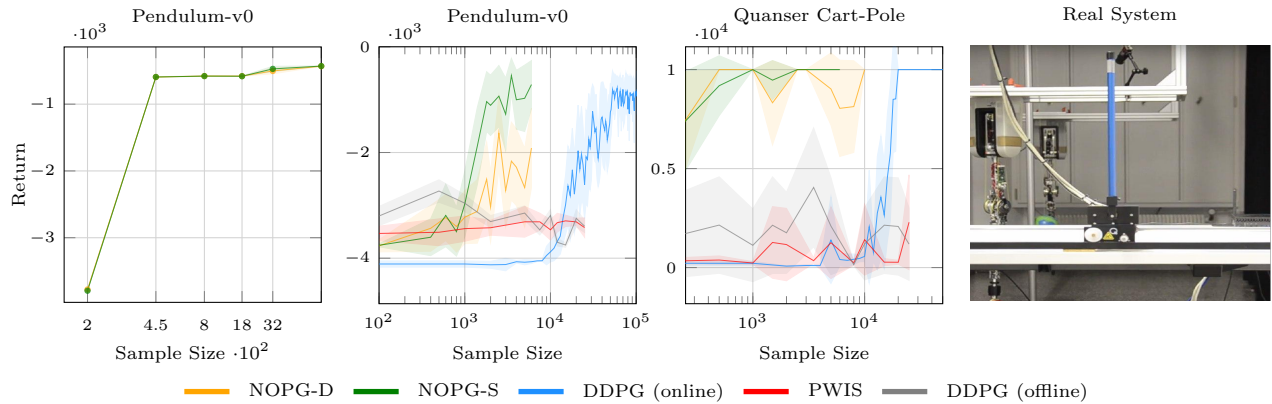

		\begin{subfigure}[t]{0.5\columnwidth}
			% This file was created by matplotlib2tikz v0.6.18.
\begin{tikzpicture}

\definecolor{color0}{rgb}{1,0.647058823529412,0}

\begin{axis}[
height=4.98cm,
width=5cm,
xmin=0, xmax=3200.0,
ymin=-3967.86533175936, ymax=-143.376848722761,
try min ticks=3,
tick align=inside,
x grid style={white!82.74509803921568!black},
y grid style={white!82.74509803921568!black},
xmajorticks=true,
ymajorticks=true,
xminorticks=true,
yminorticks=true,
xmode=log,
xmajorgrids,
ymajorgrids,
xtick={200,450, 800, 1250, 1800,3200},
xticklabels={2,4.5,8,18,32},
yticklabel style={font=\tiny},
xticklabel style={font=\tiny, yshift=-3pt},
scaled y ticks=base 10:-3,
scaled x ticks=base 10:-2,
% every x tick scale label/.style={at={(0.82, -0.21)}, anchor = south},
%every y tick scale label/.style={at={(yticklabel cs:0.1)}, anchor = north},
xlabel={Sample Size $\cdot 10^2$},
xlabel style ={font=\scriptsize, yshift=2pt},
ylabel style ={font=\scriptsize, yshift=-5pt},
ylabel={\scriptsize{Return}},
title={Pendulum-v0},
title style={font=\scriptsize},
log ticks with fixed point
]

\path [draw=color0, fill=color0, opacity=0.1] (axis cs:200,-3775.19959014973)
--(axis cs:200,-3775.18814914785)
--(axis cs:450,-591.90217816584)
--(axis cs:800,-576.783659376169)
--(axis cs:1250,-578.560411555018)
--(axis cs:1800,-459.985221504609)
--(axis cs:3200,-430.752099338165)
--(axis cs:3200,-433.108391695221)
--(axis cs:3200,-433.108391695221)
--(axis cs:1800,-545.400524422499)
--(axis cs:1250,-580.745819304394)
--(axis cs:800,-581.346848985241)
--(axis cs:450,-595.488795336325)
--(axis cs:200,-3775.19959014973)
--cycle;

\path [draw=green!50.19607843137255!black, fill=green!50.19607843137255!black, opacity=0.1] (axis cs:200,-3795.62418039938)
--(axis cs:200,-3791.22916095543)
--(axis cs:450,-590.239975072795)
--(axis cs:800,-578.148283359131)
--(axis cs:1250,-579.362788505597)
--(axis cs:1800,-433.869276677966)
--(axis cs:3200,-428.137769813099)
--(axis cs:3200,-431.767095934437)
--(axis cs:3200,-431.767095934437)
--(axis cs:1800,-510.136905039136)
--(axis cs:1250,-585.11098375442)
--(axis cs:800,-582.496690347455)
--(axis cs:450,-596.039198425476)
--(axis cs:200,-3795.62418039938)
--cycle;

\addplot [very thin, color0, opacity=0.7, mark=*, mark size=1, mark options={solid}]
table [row sep=\\]{%
	200	-3775.19386964879 \\
	450	-593.695486751082 \\
	800	-579.065254180705 \\
	1250	-579.653115429706 \\
	1800	-502.692872963554 \\
	3200	-431.930245516693 \\
};
\addplot [very thin, green!50.19607843137255!black, opacity=0.7, mark=*, mark size=1, mark options={solid}]
table [row sep=\\]{%
	200	-3793.42667067741 \\
	450	-593.139586749135 \\
	800	-580.322486853293 \\
	1250	-582.236886130008 \\
	1800	-472.003090858551 \\
	3200	-429.952432873768 \\
};
\end{axis}

\end{tikzpicture}
		\end{subfigure}\hspace{0.5em}
		\begin{subfigure}[t]{.5\columnwidth}
			\input{plots/poles/pendulumrand.tikz}
		\end{subfigure}\hspace{-0.5em}
		\begin{subfigure}[t]{.5\columnwidth}
			\input{plots/poles/cartpole.tikz}
		\end{subfigure}
		\begin{subfigure}[t]{.5\columnwidth}
			\input{plots/poles/real-cart.tikz}
		\end{subfigure}
		%	    \begin{subfigure}[t]{.5\columnwidth}
		%			\raisebox{0.87cm}{\includegraphics[height=3.5cm, width=4cm]{plots/poles/real-cart.png}}
		%		\end{subfigure}
		\vspace{-0.2cm}
		\centering
		\vspace{-0.25cm}
		\begin{tikzpicture}

\definecolor{color0}{rgb}{1,0.647058823529412,0}
\definecolor{color1}{rgb}{0.117647058823529,0.564705882352941,1}

\begin{axis}[
height=2cm,
width=8cm,
hide axis,
xmin=10,
xmax=50,
ymin=0,
ymax=0.5,
legend columns=-1,
legend entries={{NOPG-D},{NOPG-S},{DDPG (online)},{PWIS},{DDPG (offline)}},
legend style={at={(1.0,0.1)}, anchor=north, draw=none, font=\scriptsize, column sep=1ex, line width=2 pt},
]

\addlegendimage{no markers, color0}
\addlegendimage{no markers, green!50.19607843137255!black}
\addlegendimage{no markers, color1}
\addlegendimage{no markers, red}
\addlegendimage{no markers, white!50.19607843137255!black}

\end{axis}

\end{tikzpicture}
		\caption{Comparison of NOPG in its deterministic and stochastic versions to state-of-the-art algorithms on continuous control tasks: Swing-Up Pendulum with \textbf{uniform grid} sampling (left), Swing-Up Pendulum  with the \textbf{random agent} (center-left) and the Cart-Pole stabilization (center-right). The figures depict the mean and 95\% confidence interval over 10 trials. \myalg~ outperforms the baselines w.r.t the sample complexity. \textbf{Note the log-scale along the $x$-axis}. The right most picture shows the real cart-pole platform from Quanser.}
		\label{figure:comparison}
	\end{figure*}
%	\begin{table}
%		\centering
%		\begin{tabular}{|c|c|c|}
%			\hline
%			Dataset Size & NOPG-D & NOPG-S \\
%			\hline \hline
%			200 & -3555 & -3793$\pm$3  \\
%			\hline
%			450 & -593$\pm$3 & -593$\pm$3 \\
%			\hline
%			800 & -578$\pm$2  & -580$\pm$3  \\
%			\hline
%			1200 & -579$\pm$1 & -582$\pm$3 \\
%			\hline
%			1800 & -441$\pm$68 & -480$\pm$51 \\
%			\hline
%			3200 & -431$\pm$1 & -430$\pm$2 \\
%			\hline
%		\end{tabular}
%	\caption{Average return of \myalg~on the swing-up pendulum with uniform dataset.}
%	\end{table}
	\section{Empirical Evaluation}
	\label{section:experiments}
	For experimental validation we compare both the deterministic (NOPG-D) and  stochastic (NOPG-S) versions of our algorithm to G(PO)MDP with PWIS (from here on PWIS). Additionally, we compare NOPG-D to state-of-the-art deterministic off-policy algorithms DPG and DDPG. 
%	We first provide an analysis of the gradient estimation on a $2$-dimensional Linear Quadratic Regulator (LQR) problem, and then test our algorithm on the pendulum and the cart-pole tasks using different sampling techniques and compare it against the classic online version of DDPG. Eventually we test \myalg~on the mountain-car using human-demonstrations and learn a policy that surpasses the average performance of the examples provided.
%	The hyper-parameters used in the the experiments are available in the supplementary material.
	In particular we want to address the following questions:
	\begin{enumerate}
		\item How do the bias and the variance compare to PWIS and semi-gradient approaches?
		\item Does our method work in scenarios where PWIS is not applicable?
		\item How is the sample efficiency of our methods compared to state-of-the-art off-policy approaches?
	\end{enumerate}
	To answer the first question, we conduct an experiment on a $2$-dimensional LQR problem, providing a graphical representation of the gradient updates using the mentioned algorithms. We address the second question by testing our algorithm on a uniform-grid dataset (i.e. no explicit trajectories) and on a test obtained from a human demonstrator. 
	To test the sample efficiency we compare our methods against DDPG, offline DDPG, and PWIS on the swing-up pendulum and on the cart-pole stabilization\footnote{The code of \myalg~is available at \url{https://github.com/jacarvalho/nopg}.}.
	The supplementary provides details of all hyper-parameters used, an implementation of NOPG and video of the final policy executed on a real cart-pole system.

	\subsection{Gradient Direction with LQR}
		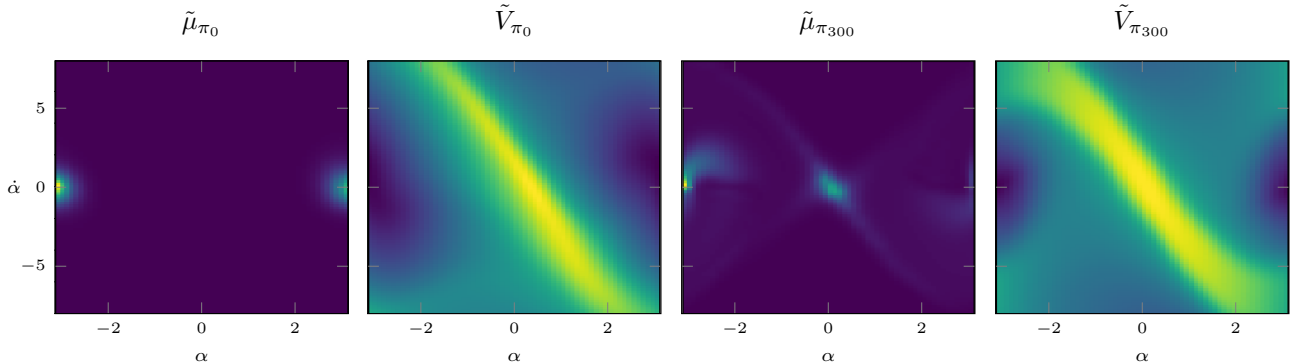
\begin{figure*}
		\centering
		\hspace{-1.0em}
		\begin{subfigure}[t]{0.5\columnwidth}
			\begin{tikzpicture}
    \begin{axis}[
        axis on top,% ----
        width=3.9cm,
        scale only axis,
        enlargelimits=false,
        xmin=-3.14,
        xmax=3.14,
        ymin=-8,
        ymax=8,
        title={$\tilde{\mu}_{\pi_0}$},
        xlabel={$\alpha$},
        ylabel={$\dot{\alpha}$},
        xlabel style ={font=\scriptsize},
        ylabel style ={font=\scriptsize, rotate=-90, xshift=7pt},
        yticklabel style={font=\tiny},
        xticklabel style={font=\tiny}
        ]
      \addplot[thick,blue] graphics[xmin=-3.14,ymin=-8,xmax=3.14,ymax=8] {plots/poles/mu0new.pdf};
    \end{axis}
  \end{tikzpicture}
		\end{subfigure}\hspace{1.2em}
		\begin{subfigure}[t]{0.5\columnwidth}
			\begin{tikzpicture}
    \begin{axis}[
        axis on top,% ----
        width=3.9cm,
        scale only axis,
        enlargelimits=false,
        xmin=-3.14,
        xmax=3.14,
        ymin=-8,
        ymax=8,
        title={$\tilde{V}_{\pi_0}$},
        xlabel={$\alpha$},
        xlabel style ={font=\scriptsize},
        yticklabel =\empty,
        xticklabel style={font=\tiny}
        ]
      \addplot[thick,blue] graphics[xmin=-3.14,ymin=-8,xmax=3.14,ymax=8] {plots/poles/v0_new.pdf};
    \end{axis}
  \end{tikzpicture}
		\end{subfigure}\hspace{-0.2em}
		\begin{subfigure}[t]{0.5\columnwidth}
			\begin{tikzpicture}
    \begin{axis}[
        axis on top,% ----
        width=3.9cm,
        scale only axis,
        enlargelimits=false,
        xmin=-3.14,
        xmax=3.14,
        ymin=-8,
        ymax=8,
        title={$\tilde{\mu}_{\pi_{300}}$},
        xlabel={$\alpha$},
		xlabel style ={font=\scriptsize},
		yticklabel =\empty,
		xticklabel style={font=\tiny}
        ]
      \addplot[thick,blue] graphics[xmin=-3.14,ymin=-8,xmax=3.14,ymax=8] {plots/poles/mu1_new.pdf};
    \end{axis}
  \end{tikzpicture}
		\end{subfigure}\hspace{-0.2em}
		\begin{subfigure}[t]{0.5\columnwidth}
			\begin{tikzpicture}
    \begin{axis}[
        axis on top,% ----
        width=3.9cm,
        scale only axis,
        enlargelimits=false,
        xmin=-3.14,
        xmax=3.14,
        ymin=-8,
        ymax=8,
        title={$\tilde{V}_{\pi_{300}}$},
        xlabel={$\alpha$},
		xlabel style ={font=\scriptsize},
		yticklabel =\empty,
		xticklabel style={font=\tiny}
        ]
      \addplot[thick,blue] graphics[xmin=-3.14,ymin=-8,xmax=3.14,ymax=8] {plots/poles/v1_new.pdf};
    \end{axis}
  \end{tikzpicture}
		\end{subfigure}\vspace{-1em}
		\caption{A phase portrait of the state distribution $\tilde{\mu}_{\pi}$ and value function $\tilde{V}_{\pi}$ estimated in the swing-up pendulum task with NOPG-D. Green corresponds to higher values. The two leftmost figures show the estimates before any policy improvement, while the two rightmost show them after $300$ offline updates of \myalg-D. Notice that the algorithm finds a very good approximation of the optimal value function and is able to predict that the system will reach the goal state ($(\alpha, \dot{\alpha}) = (0, 0) $).}
		\label{figure:muv}
		\vskip -0.2in
		\end{figure*}
%	\vspace{-2em}
	We qualitatively demonstrate how the different gradient estimates work on a simple $2$-dimensional problem. For this purpose we choose a linear-quadratic regulator setup, and use a linear policy encoded by the diagonal matrix $K = [[k_1,0], [0,k_2]]$. Figure~\ref{figure:gradient} illustrates experiments with deterministic and stochastic policies. In the experiment with deterministic policies we evaluate the performance of \myalg-D and offline-DPG over $5$ datasets, each containing $100$ trajectories of length $30$.
	The experiment with stochastic policies compares the gradient estimates of \myalg-S and PWIS. Given that PWIS requires a single stochastic policy, we generated $5$ datasets with $100$ trajectories of length $30$ from the interactions of a Gaussian policy with the environment.
	The results in Figure~\ref{figure:gradient} show that PWIS suffers from high variance while DPG offers a biased estimate, which is consistent with our initial theoretical hypothesis. 
	Moreover, it is interesting to observe how the error in DPG's biased estimate compounds after every iteration as the algorithm moves away from the initial ``on-policy'' region in the vicinity of the behavioral policy. \myalg~on the other hand exhibits a more accurate gradient estimate in the off-policy region and with smaller variance when compared to PWIS.
	
	\subsection{Swing-Up Pendulum and Cart-Pole}
	The under-powered pendulum and the cart-pole are two classical control tasks often used in RL for empirical analysis. We use the OpenAI Gym framework \citep{brockman_openai_2016} for a pendulum simulation, and implement another environment that simulates the dynamics of a real cart-pole built by Quanser \footnote{\url{https://www.quanser.com/products/linear-servo-base-unit-inverted-pendulum}}
	
	\paragraph{Uniform Grid.} 
	In this experiment we analyze the performance of \myalg~under a uniformly sampled dataset, since, as the theory suggests, this scenario should yield the least biased estimate of NOPG. We generate datasets from a grid over the state-action space of the pendulum environment with different granularities. We test our algorithm by optimizing a policy encoded with a neural-network for a fixed amount of iterations. The policy is composed of a single hidden layer of $50$ neurons with ReLU activations. This configuration is fixed across all the different experiments and algorithms for the remainder of this document.
	The resulting policy is evaluated on trajectories of $500$ steps starting from the bottom position.
	The leftmost plot in Figure~\ref{figure:comparison}, depicts the performance against different dataset sizes, showing that \myalg~is able to solve the task with $450$ samples. Figure~\ref{figure:muv} is an example of the value function and state distribution estimates of \myalg-D at the beginning and after $300$ optimization steps. The ability to predict the state-distribution is particularly interesting for robotics, as it is possible to predict in advance whether the policy will move towards dangerous states. Note that this experiment is not applicable to PWIS, as it does not admit non-trajectory-based data.
	
	\paragraph{Random Agent.}
	In contrast to the uniform grid experiment, here we collect the datasets using trajectories from a random agent in the pendulum and the cart-pole environments. 
	In the pendulum task, the trajectories are generated starting from the up-right position and applying a policy composed of a mixture of Gaussians. The policies are evaluated starting from the bottom position with an episode length of $500$ steps. The datasets used in the cart-pole experiments are collected using a uniform policy starting from the upright position until the end of the episode, which occurs when the absolute value of the  angle $\theta$ surpasses $3\deg$. The optimization policy is evaluated for $10^4$ steps. The reward is $r_t = \cos \theta_t$. Since $\theta$ is defined as $0$ in the top-right position, a return of $10^4$ indicates an optimal policy behavior. 
	
	We analyze the sample efficiency by testing \myalg, PWIS  and DDPG in an offline fashion with pre-collected samples, on different number of trajectories. In addition, we provide the learning curve with the classical online DDPG using the OpenAI Baselines implementation \citep{dhariwal_openai_2017-1}. 
	
	We stress that, since offline DDPG and PWIS show an unstable learning curve, we always report the \textsl{best evaluation} obtained during the learning process, while with \myalg~we report the last evaluation.
	The two center plots in Figure~\ref{figure:comparison} highlight that our algorithm has superior sample efficiency by more than one order of magnitude (note the log-scale on the x-axis).
	
	To validate the resulting policy learned in simulation, we apply the final learned controller on a real Quanser cart-pole, and observe a successful stabilizing behavior as can be seen in the supplementary video.
		
	\subsection{Mountain Car with Human Demonstrations} 
		\begin{table*}[t]
		\centering
		%		\begin{tabular}{|l|c|c|c|}
		%			\hline
		%			& Semi-Gradient & PWIS &  NOPG \\
		%			\hline\hline 
		%			Human Demonstration & \cmark & \xmark &  \cmark \\
		%			\hline 
		%			Unstructured Datasets & \cmark & \xmark &  \cmark \\
		%			\hline 
		%			Deterministic Policies & \cmark & \xmark &  \cmark \\
		%			\hline 
		%			Bias & \thigh & \tlow &  \tlow \\
		%			\hline 
		%			Variance & \tlow & \thigh &  \tlow \\
		%%			\hline 
		%%			Infinite Horizon & \cmark & \cmark & \xmark & \cmark \\
		%%			\hline 
		%%			Multimodal Transition & \cmark & \cmark & \xmark & \cmark \\
		%%			\hline 
		%%			Scalable w.r.t. Dimensions & \cmark & \xmark & \xmark & \cmark \\
		%			\hline
		%		\end{tabular}
		\begin{tabular}{lccccc}
			\hline
			& Human Demonstration & Unstructured Dataset & Deterministic Policies & Bias & Variance \\
			\hline
			Semi-Gradient & \cmark & \cmark &  \cmark & \thigh & \tlow \\
			PWIS & \xmark & \xmark &  \xmark  & \tlow & \thigh \\
			NOPG& \cmark & \cmark &  \cmark  &\tlow & \tlow \\
			%			\hline 
			%			Infinite Horizon & \cmark & \cmark & \xmark & \cmark \\
			%			\hline 
			%			Multimodal Transition & \cmark & \cmark & \xmark & \cmark \\
			%			\hline 
			%			Scalable w.r.t. Dimensions & \cmark & \xmark & \xmark & \cmark \\
			\hline
		\end{tabular}
		\caption{Applicability of off-policy algorithms. Our algorithm is applicable to a wider range of tasks in contrast to state-of-the-art techniques. NOPG is able to deal with human demonstrations and unstructured datasets by using either a stochastic or a deterministic policy, all while exhibiting lower bias and variance than its competitors. \label{table:applicability}}
	\end{table*}
	
\begin{figure}[t]
		\centering
		\begin{subfigure}{0.49\columnwidth}
			%\begin{tikzpicture}
%\node (cartpole) at (0,0) {\includegraphics[height=6.2cm, width=8.0cm]{plots/Mountain/mountaincar_samplesize.pdf}};
%\node [above=-0.5cm of cartpole.north] {\textsuperscript{Mountain Car}};
%%\node [below=.45cm of cartpole.south] {\textsuperscript{}};
%\end{tikzpicture}

% This file was created by matplotlib2tikz v0.6.18.
\begin{tikzpicture}

\definecolor{color0}{rgb}{1,0.647058823529412,0}

\begin{axis}[
height=4.6cm,
width=4.8cm,
legend cell align={left},
legend columns=1,
legend entries={{NOPG-D},{NOPG-S},{Human baseline}},
legend style={at={(0.5,-0.5)}, anchor=center, draw=none, font=\scriptsize, line width=2pt},
tick pos=left,
x grid style={white!82.74509803921568!black},
xlabel={Number of trajectories},
xmajorgrids,
yticklabel style={font=\tiny},
xticklabel style={font=\tiny},
xmin=1, xmax=10,
y grid style={white!82.74509803921568!black},
ylabel={Return},
ymajorgrids,
ymin=-452.243393203803, ymax=-53.0887427201282,
ytick={-500,-400,-300,-200,-100},
yticklabels={−5,−4,−3,−2,−1},
scaled y ticks=base 10:-2,
xlabel style ={font=\scriptsize},
ylabel style ={font=\scriptsize, yshift=-5.0pt},
title={Mountain Car},
title style={font=\scriptsize}
]
\addlegendimage{no markers, color0}
\addlegendimage{no markers, green!50.19607843137255!black}
\addlegendimage{no markers, black, dashed}
\path [draw=color0, fill=color0, opacity=0.1] (axis cs:1,-390.043726984244)
--(axis cs:1,-158.356273015756)
--(axis cs:2,-83.5504952505809)
--(axis cs:3,-100.087436344246)
--(axis cs:4,-94.0036470748486)
--(axis cs:5,-71.2321359239316)
--(axis cs:6,-75.2454775399103)
--(axis cs:7,-76.8341531201946)
--(axis cs:8,-72.0814498628687)
--(axis cs:9,-95.216)
--(axis cs:10,-78.8726402026378)
--(axis cs:10,-93.5273597973622)
--(axis cs:10,-93.5273597973622)
--(axis cs:9,-96.784)
--(axis cs:8,-85.9185501371313)
--(axis cs:7,-95.3658468798054)
--(axis cs:6,-90.3545224600897)
--(axis cs:5,-82.7678640760684)
--(axis cs:4,-119.396352925151)
--(axis cs:3,-128.712563655754)
--(axis cs:2,-118.449504749419)
--(axis cs:1,-390.043726984244)
--cycle;

\path [draw=green!50.19607843137255!black, fill=green!50.19607843137255!black, opacity=0.1] (axis cs:1,-410.956186102897)
--(axis cs:1,-155.488258341548)
--(axis cs:2,-81.3742096464664)
--(axis cs:3,-94.0336100314196)
--(axis cs:4,-82.7473071310021)
--(axis cs:5,-82.7377886289309)
--(axis cs:6,-78.8041515604406)
--(axis cs:7,-76.4060938206283)
--(axis cs:8,-85.089922003053)
--(axis cs:9,-76.0101712106379)
--(axis cs:10,-74.5548894076772)
--(axis cs:10,-92.3339994812117)
--(axis cs:10,-92.3339994812117)
--(axis cs:9,-93.3231621226955)
--(axis cs:8,-98.2434113302803)
--(axis cs:7,-93.3439061793717)
--(axis cs:6,-93.4458484395594)
--(axis cs:5,-115.706655815514)
--(axis cs:4,-110.363803980109)
--(axis cs:3,-102.21638996858)
--(axis cs:2,-109.514679242423)
--(axis cs:1,-410.956186102897)
--cycle;

\addplot [very thin, color0, opacity=0.7, mark=*, mark size=1, mark options={solid}]
table [row sep=\\]{%
	1	-274.2 \\
	2	-101 \\
	3	-114.4 \\
	4	-106.7 \\
	5	-77 \\
	6	-82.8 \\
	7	-86.1 \\
	8	-79 \\
	9	-96 \\
	10	-86.2 \\
};
\addplot [very thin, green!50.19607843137255!black, opacity=0.7, mark=*, mark size=1, mark options={solid}]
table [row sep=\\]{%
	1	-283.222222222222 \\
	2	-95.4444444444444 \\
	3	-98.125 \\
	4	-96.5555555555556 \\
	5	-99.2222222222222 \\
	6	-86.125 \\
	7	-84.875 \\
	8	-91.6666666666667 \\
	9	-84.6666666666667 \\
	10	-83.4444444444444 \\
};
\addplot [black, opacity=1., dashed]
table [row sep=\\]{%
	0	-434.1 \\
	10	-434.1 \\
};
\path [draw=black, fill opacity=0] (axis cs:0,-452.243393203803)
--(axis cs:0,-53.0887427201282);

\path [draw=black, fill opacity=0] (axis cs:1,-452.243393203803)
--(axis cs:1,-53.0887427201282);

\path [draw=black, fill opacity=0] (axis cs:1,0)
--(axis cs:10,0);

\path [draw=black, fill opacity=0] (axis cs:1,1)
--(axis cs:10,1);
\end{axis}
\end{tikzpicture}
		\end{subfigure}\hspace{-0.1em}
		\begin{subfigure}{0.49\columnwidth}
			\input{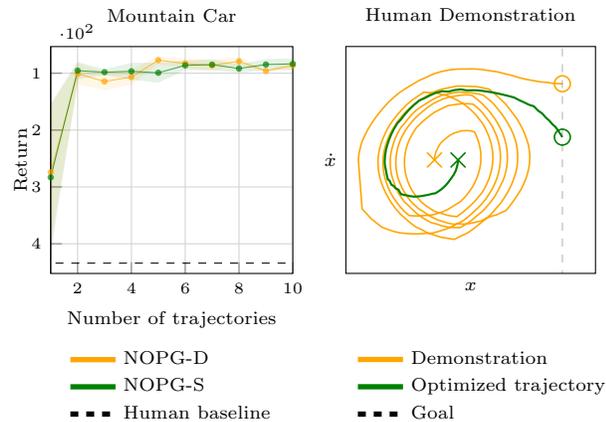}
		\end{subfigure}\vspace{-1em}
		\caption{With a small amount of data NOPG is able to reach a policy that surpasses the human demonstrator (dashed line) in the mountain car environment. Depicted are the mean and 95\% confidence over 10 trials (left). An example of a human-demonstrated trajectory and the relative optimized version obtained with \myalg~(right). Although the human trajectories in the dataset are suboptimal, \myalg~converges to an optimal solution (right).}
		\label{figure:mountain}
	\end{figure}

	%\vspace{-1em}
	In robotics, learning from human demonstrations is crucial in order to obtain better sample efficiency and to avoid dangerous policies. This experiment is designed to showcase the ability of our algorithm to deal with such demonstrations without the need for explicit knowledge of the underlying behavioral policy. The experiment is executed in a completely offline fashion after collecting the human dataset, i.e., without any further interaction with the environment. This setting is different from the classical imitation learning and subsequent optimization \citep{kober_policy_2009}.
	As an environment we choose the continuous mountain car task from OpenAI. We provide $10$ demonstrations recorded by a human operator and assigned a reward of $-1$ to every step. A demonstration ends when the human operator surpasses the limit of $500$ steps, or arrives at the goal position. The human operator explicitly provides sub-optimal trajectories, as we are interested in analyzing whether \myalg~is able to take advantage of the human demonstrations to learn a better policy than that of the human, without any further interaction with the environment. To obtain a sample analysis, we evaluate NOPG on randomly selected sub-sets of the trajectories from the human demonstrations. Figure~\ref{figure:mountain} shows the average performance as a function of the number of demonstrations as well as an example of a human-demonstrated trajectory.
	Notice that both \myalg-S and \myalg-D manage to learn a policy that surpasses the human operator's performance and reach the optimal policy with two demonstrated trajectories.
	
%	Our method, unlike PWIS, does not require the knowledge of the behavioral policy nor the concept of trajectory. Moreover the avoidance of importance sampling and the usage of the critic, provides a lower variance in the estimation of the gradient. On the other hand, our estimation suffers from the intrinsic bias of the value function approximation, like other actor/critic methods, but our full-gradient estimate eliminates a further source of bias which is present in semi-gradient approaches. Table~\ref{table:applicability} shows concisely the benefits of our methods in contrast to PWIS and semi-gradient estimation.

	\section{Conclusion and Future Work}
	We proposed a novel off-policy policy gradient method that is based on a nonparametric Bellman equation and provides a full-gradient estimate that can be computed in closed-form. Our approach avoids the issues of pathwise importance sampling and semi-gradient methods. More explicitly, the full-gradient estimate is less biased than the semi-gradient, and exhibits lower variance that the gradient estimates of importance-sampling-based approaches. Moreover, our algorithm enables learning from human demonstrations and non-trajectory-based datasets. An overview highlighting the pros and cons of all mentioned approaches is given in Table~\ref{table:applicability}.
	
	To support our argument and findings we provide both a theoretical and empirical analysis conducted in different scenarios and compare to state-of-the-art off-policy algorithms. Our theoretical analysis provides a bound on the estimation bias and highlights the impact of different factors. The empirical analysis shows that our method succeeded in learning near-optimal policies in off-policy settings where semi-gradient approaches fail. Furthermore, our approach was able to leverage unstructured datasets and human demonstrations, two scenarios where importance sampling techniques are not applicable.
	
	The accurate gradient estimate delivered by our algorithm results in dramatically overall lower sample complexity when compared to state-of-the-art off-policy policy gradients. By relying on nonparametric statistics, we sacrifice scalability for higher sample efficiency and safety, bringing reinforcement learning one step closer to real world applications and robotics.
	
	Future research will concentrate on extending our approach to parametric models to address scalability, exploring the possibility of using \myalg~ in a Bayesian framework in order to deal with the problems of uncertainty and model bias, and on enabling a principled online exploration and informative data collection.
	
	\section{Acknowledgment}
	The research is financially supported by the Bosch-Forschungsstiftung program and the European Union’s Horizon 2020 research and innovation program under grant agreement \#640554 (SKILLS4ROBOTS).
	
	\bibliography{zotero}
	\bibliographystyle{aistats2020}
	
	\newpage
	\appendix
	\onecolumn
%\begin{adjustwidth}{-75pt}{-60pt}

{\huge{Supplementary Material}}

\section{The Nonparametric Bellman Equation}
This section contains the proofs of Theorem~\ref{theorem:fixedpoint} and Theorem~\ref{theorem:ultimate}. \\
\begin{proposition}
	In the limit of infinite samples the NPBE defined in Definition~\ref{definition:npbe} with a data-set $\lim_{n\to \infty}D_n$ collected under distribution $\beta$ on the state-action space and MDP $\mathcal{M}$ converges to
	\begin{eqnarray}
		\approxx{V}_{\pi}(\svec)  & = &  \int_{\Sset \times \Aset} \varepsilon_{\pi}(\svec,\zvec, \bvec) \bigg( R_{\zvec, \bvec} + \gamma \int_{\Sset}\approxx{V}_{\pi}(\svec')\phi(\svec', \zvec_{\zvec, \bvec}')\de \svec'  \bigg) \beta(\zvec, \bvec) \de \zvec \de \bvec, \nonumber \\
		& & \text{with} \quad R_{\zvec, \bvec} \sim R(\zvec, \bvec) \quad \forall(\zvec, \bvec) \in \Sset \times \Aset, \nonumber \\
		& & \text{with} \quad \zvec_{\zvec, \bvec}' \sim P(\cdot |\zvec, \bvec) \quad \forall(\zvec, \bvec) \in \Sset \times \Aset. \label{equation:infnpbe}
	\end{eqnarray}
	and
	\begin{equation}
		\begin{cases}
			\varepsilon_{\pi}(\svec, \zvec, \bvec) \defeq \bigintsss_{\Aset} \displaystyle \frac{\psi(\svec, \zvec)\varphi(\avec, \bvec)}{\int_{\Sset, \Aset}\psi(\svec, \zvec)\varphi(\avec, \bvec)\beta(\zvec, \bvec) \de \zvec \de \bvec} \pi(\avec | \svec) \de \avec & \text{if $\pi$ is stochastic}, \\
			\\
			\varepsilon^{\pi}_i(\svec)  \defeq \displaystyle \frac{\psi(\svec, \zvec)\varphi(\pi(\svec),\bvec)}{\int_{\Sset, \Aset} \psi(\svec, \zvec)\varphi(\pi(\svec), \bvec)\beta(\zvec, \bvec) \de \zvec \de \bvec}                                           & \text{otherwise.}
		\end{cases}\nonumber
	\end{equation}
	\begin{proof}
		\begin{eqnarray}
			\approxx{V}_{\pi}(\svec) & = & \lim_{n \to \infty} \int_{\Aset}\frac{\sum_{i=1}^n\psi_i(\svec)\varphi_i(\avec)\bigg(r_i + \gamma \int_{\Sset}\phi_i(\svec')\approxx{V}_{\pi}(\svec')\de \svec\bigg)}{\sum_{i=1}^n\psi_j(\svec)\varphi_j(\avec)} \pi(\avec | \svec) \de \avec \nonumber \\
			& = & \int_{\Aset}\frac{\lim_{n \to \infty}\frac{1}{n}\sum_{i=1}^n\psi_i(\svec)\varphi_i(\avec)\bigg(r_i + \gamma \int_{\Sset}\phi_i(\svec')\approxx{V}_{\pi}(\svec')\de \svec\bigg)}{\lim_{n \to \infty} \frac{1}{n}\sum_{i=1}^n\psi_j(\svec)\varphi_j(\avec)} \pi(\avec | \svec) \de \avec \nonumber \\
			& = & \int_{\Aset}\frac{\int_{\Sset \times \Aset}\psi(\svec, \zvec)\varphi(\avec, \bvec)\bigg( R(\zvec, \bvec) + \gamma \int_{\Sset}\phi(\svec', \zvec')p(\zvec'|\bvec, \zvec)\approxx{V}_{\pi}(\svec')\de \svec  \bigg) \beta(\zvec, \bvec) \de \zvec \de \bvec}{\int_{\Sset \times \Aset}\psi(\svec, \zvec)\varphi(\avec, \bvec)\beta(\zvec, \bvec) \de \zvec \de \bvec} \pi(\avec | \svec) \de \avec. \nonumber
		\end{eqnarray}
		Analogously we can derive the deterministic policy case.
	\end{proof}
\end{proposition}
\begin{proposition}
We want to show that, if a solution $\hat{V}_\pi(\state)$ exists, it is bounded by $|\hat{V}_\pi(\state)| \leq R_{\max}/(1-\gamma)$, where $R_{\max} = \max_i |r_i|$.
\begin{proof}
	Suppose by absurd that exists a state $\zvec \in \Sset$ such that 
	$|\hat{V}_\pi(\zvec)| = R_{\max}/(1-\gamma) + \epsilon$ with $\epsilon \in \mathbb{R}^+$. Then,
	
	\begin{equation}
	\frac{R_{\max}}{1-\gamma} + \epsilon = \bm{\varepsilon}^\intercal(\zvec) \rvec + \gamma \bm{\varepsilon}^\intercal(\zvec) \int_\Sset \phivec(\svec')\hat{V}_\pi(\svec')\de \svec' .
	\end{equation}
	
	Since, by definition, the equation must be correctly fulfilled, we notice that $|\varepsilon_\pi^\intercal(\zvec)| \leq R_{\max}$, since $\varepsilon_\pi(\zvec)$ is a stochastic vector, therefore
	
	\begin{eqnarray}
	\left| \frac{R_{\max}}{1-\gamma} + \epsilon - \gamma \bm{\varepsilon}^\intercal(\zvec) \int_\Sset \phivec(\svec')\hat{V}_\pi(\svec')\de \svec' \right| \leq R_{\max} . \label{eq:passage:reward_bounding}
	\end{eqnarray}
	
	However, 
	
	\begin{eqnarray}
	\frac{R_{\max}}{1-\gamma} + \epsilon - \gamma \bm{\varepsilon}^\intercal(\zvec) \int_\Sset \phivec(\svec')\hat{V}_\pi(\svec')\de \svec' & \geq &  \frac{R_{\max}}{1-\gamma} + \epsilon - \gamma\left( \frac{R_{\max}}{1-\gamma} +\epsilon \right)\nonumber \\
	& \geq &  R_{\max} + \gamma \epsilon \nonumber 
	\end{eqnarray}
	this is in contraddiction with \ref{eq:passage:reward_bounding}.
\end{proof}
\end{proposition}

\begin{proposition}
	\label{proposition:uniqueness}
	If $R$ is bounded by $R_{\text{max}}$ and if $f^*:\Sset \to \mathbb{R}$ satisfies the NPBE, then there is no other function $f:\Sset \to \mathbb{R}$ for which $\exists \zvec \in \Sset$ and $|f^*(\zvec) - f(\zvec)| > 0$.
	\begin{proof}
		Suppose, by absurd assumption, that a function $g:\Sset \to \mathbb{R}$ exists such that $f(\svec) + g(\svec)$ satisfies Equation~\eqref{equation:infnpbe} for every $\svec \in \Sset$ and a constant $G \in \mathbb{R}^+$ exists for which $|g(\zvec)| > G$.
		Note that the existence of $f:\Sset\to\mathbb{R}$ as a solution for the NPBE implies the existence of
		\begin{equation}
			\int_{\Sset}\bm{\varepsilon}_{\pi}^T(\svec)\phivec(\svec')f^*(\svec ' )\de   \svec ' \in \mathbb{R}, \label{equation:existenceint1}
		\end{equation}
		and similarly, the existence of $f(\svec) \in \mathbb{R}$ with $f(\svec) = f^*(\svec) + g(\svec)$ as a solution of the NPBE implies that
		\begin{equation}
			\int_{\Sset}\bm{\varepsilon}_{\pi}^T(\svec)\phivec(\svec')f^*(\svec ' ) + g(\svec')\de   \svec ' \in \mathbb{R}. \label{equation:existenceint2}
		\end{equation}
		Note that the existence of the integral in Equations~\eqref{equation:existenceint1} and \eqref{equation:existenceint2} implies
		\begin{equation}
			\int_{\Sset}\bm{\varepsilon}_{\pi}^T(\svec)\phivec(\svec') g(\svec')\de   \svec ' \in \mathbb{R}. \label{equation:existenceint3}
		\end{equation}
		Note that
		\begin{eqnarray}
			|f^*(\svec) - f(\svec)| & = & \bigg| f^*(\svec) - \bm{\varepsilon}_{\pi}^T(\svec) \bigg(\rvec + \gamma \int_{\Sset} \phivec(\svec') \big(f(\svec ') + g(\svec ')\big)  \de \nonumber \svec'\bigg) \bigg| \nonumber \\
			& = & \bigg| \bm{\varepsilon}_{\pi}^T(\svec) \bigg(\rvec + \gamma \int_{\Sset} \phivec(\svec')  g(\svec' )  \de \nonumber \svec' \bigg)- \bm{\varepsilon}_{\pi}^T(\svec) \bigg(\rvec + \gamma \int_{\Sset} \phivec(\svec') \big(f^*(\svec ') + g(\svec ')\big)  \de \nonumber \svec'\bigg) \bigg| \nonumber \\
			& = & \gamma \bigg| \bm{\varepsilon}_{\pi}^T(\svec)   \int_{\Sset} \phivec(\svec') g(\svec ')  \de \nonumber \svec'\bigg| \nonumber \\
			\implies |g(\svec)| & = & \gamma \bigg| \bm{\varepsilon}_{\pi}^T(\svec)   \int_{\Sset} \phivec(\svec') g(\svec ' )  \de \nonumber \svec'\bigg|. \nonumber
		\end{eqnarray}
		Using Jensen's inequality
		\begin{eqnarray}
			|g(\svec)|& \leq & \gamma \bm{\varepsilon}_{\pi}^T(\svec)   \int_{\Sset} \phivec(\svec')  |g(\svec ') |  \de \nonumber \svec'. \nonumber
		\end{eqnarray}
		Note that since both $f^*$ and $f$ are bounded by $\frac{R_{\text{max}}}{1-\gamma}$ then $|g(\svec)| \leq \frac{2R_{\text{max}}}{1-\gamma}$, thus
		\begin{eqnarray}
			|g(\svec)|& \leq & \gamma \bm{\varepsilon}_{\pi}^T(\svec)   \int_{\Sset} \phivec(\svec')  |g(\svec ') |  \de  \svec' \label{equation:recursion} \\
			& \leq & \gamma  2\frac{R_{\text{max}}}{1-\gamma} \bm{\varepsilon}_{\pi}^T(\svec)   \int_{\Sset} \phivec(\svec')    \de \nonumber \svec' \nonumber \\
			& = & \gamma  2\frac{R_{\text{max}}}{1-\gamma}  \nonumber \\
			\implies |g(\svec)|& \leq & \gamma \frac{2R_{\text{max}}}{1-\gamma} \nonumber \\
			\implies |g(\svec)|& \leq & \gamma^2 \frac{2R_{\text{max}}}{1-\gamma} \quad \quad \text{using \eqref{equation:recursion}}\nonumber \\
			\implies |g(\svec)|& \leq & \gamma^3 \frac{2R_{\text{max}}}{1-\gamma} \quad \quad \text{using \eqref{equation:recursion}}\nonumber \\
			\dots \nonumber \\
			\implies |g(\svec)|& \leq &  0, \nonumber
		\end{eqnarray}
		which is in clear disagreement with the assumption made. Again here a similar procedure shows the same result for the infinite case.
	\end{proof}
\end{proposition}

\paragraph{Proof of Theorem~\ref{theorem:fixedpoint}}
\begin{proof}
	Saying that $\approxx{V}_{\pi}^*$ is a solution for Equation~\eqref{equation:infnpbe} is equivalent to saying
	\begin{equation}
		\approxx{V}_{\pi}^*(\svec) - \bm{\varepsilon}^{\pi}(\svec) \bigg(\rvec + \gamma \int_{\Sset} \phivec(\svec')\approxx{V}_{\pi}^*(\svec')   \de \svec'\bigg) = 0 \quad \quad \forall \svec \in \Sset. \nonumber
	\end{equation}
	We can verify that by simple algebraic manipulation
	\begin{eqnarray}
		& & \approxx{V}_{\pi}^*(\svec) - \bm{\varepsilon}_{\pi}^T(\svec) \bigg(\rvec + \gamma \int_{\Sset} \phivec(\svec')\approxx{V}_{\pi}^*(\svec')   \de \svec'\bigg) \nonumber \\
		& = & \bm{\varepsilon}_{\pi}^T(\svec) \bm{\Lambda}_{\pi}^{-1} \rvec - \bm{\varepsilon}^{\pi}(\svec) \bigg(\rvec + \gamma \int_{\Sset} \phivec(\svec')\bm{\varepsilon}_{\pi}^T(\svec') \bm{\Lambda}_{\pi}^{-1}\rvec   \de \svec'\bigg) \nonumber \\
		& = & \bm{\varepsilon}_{\pi}^T(\svec)\bigg(\bm{\Lambda}_{\pi}^{-1} \rvec - \rvec - \gamma \int_{\Sset} \phivec(\svec')\bm{\varepsilon}_{\pi}^T(\svec') \bm{\Lambda}_{\pi}^{-1}\rvec   \de \svec'\bigg) \nonumber \\
		& = & \bm{\varepsilon}_{\pi}^T(\svec)\Bigg(\bigg(I- \gamma \int_{\Sset} \phivec(\svec')\bm{\varepsilon}_{\pi}^T(\svec')   \de \svec'\bigg)\bm{\Lambda}_{\pi}^{-1}\rvec - \rvec   \Bigg) \nonumber \\
		& = & \bm{\varepsilon}_{\pi}^T(\svec)\Bigg(\bm{\Lambda}_{\pi}\bm{\Lambda}_{\pi}^{-1}\rvec - \rvec   \Bigg) \nonumber \\
		& = & 0.
	\end{eqnarray}
	Since equation~\eqref{equation:infnpbe} has (at least) one solution, Proposition~\ref{proposition:uniqueness} guarantees that the solution ($\approxx{V}_{\pi}^*$) is unique.
\end{proof}

\paragraph{Proof of Theorem~\ref{theorem:ultimate}.}
\begin{proof}
	We perform the derivation for the stochastic policy, however the same derivation applies for the deterministic case almost identically.
	Expanding $\big|\EV_{D}[\overline{V}_D(\svec)] - V^*(\svec)\big|$ using the NPBE and the classic Bellman equation,
	\begin{eqnarray}
		\big|\EV_{D}[\overline{V}_D(\svec)] - V^*(\svec)\big| &= &\Bigg| \EV_D\bigg[\int_{\Sset \times \Aset} \varepsilon_{\pi}(\svec,\zvec, \bvec) \bigg( R_{\zvec, \bvec} + \gamma \int_{\Sset}V_D(\svec')\phi(\svec', \zvec_{\zvec, \bvec}')\de \svec  \bigg) \beta(\zvec, \bvec) \de \zvec \de \bvec \nonumber \bigg] \nonumber \\
		& & - \int_{\Aset} \bigg(\overline{R}(\svec,\avec) + \gamma \int_{\Sset} V^*(\svec')p(\svec'|\svec, \avec)\de \svec' \bigg)\pi(\avec|\svec)\de \avec\Bigg|. \label{equation:expand}
	\end{eqnarray}
	As can be easily verified, $\varepsilon_{\pi}(\svec,\zvec,\bvec)\beta(\zvec,\bvec)$ is a density distribution over $\zvec, \bvec$. Hence Equation~\eqref{equation:expand} can be rewritten
	\begin{eqnarray}
		& & \Bigg|\EV_D\bigg[\int_{\Sset \times \Aset} \varepsilon_{\pi}(\svec,\zvec, \bvec) \bigg( R_{\zvec, \bvec} + \gamma \int_{\Sset}V_D(\svec')\phi(\svec', \zvec_{\zvec, \bvec}')\de \svec'  \bigg) \beta(\zvec, \bvec) \de \zvec \de \bvec \nonumber \bigg] \nonumber \\
		& & - \int_{\Aset} \bigg(\overline{R}(\svec,\avec)  + \gamma \int_{\Sset} V^*(\svec')p(\svec'|\svec, \avec)\de \svec' \bigg)\pi(\avec|\svec)\de \avec\Bigg| \nonumber \\
		& = & \Bigg|\EV_D\bigg[ \int_{\Aset}\frac{\int_{\Sset \times \Aset}\psi(\svec, \zvec)\varphi(\avec, \bvec)\big( R_{\zvec, \bvec}  - \overline{R}(\svec,\avec)\big)\beta(\zvec, \bvec)\de \zvec \de \bvec}{\int_{\Sset, \Aset}\psi(\svec, \zvec)\varphi(\avec, \bvec)\beta(\zvec, \bvec) \de \zvec \de \bvec}  \pi(\avec | \svec) \de \avec \bigg]\nonumber \\
		&  +  & \gamma  \int_{\Aset}\EV_D\bigg[\frac{\int_{\Sset \times \Aset}\psi(\svec, \zvec)\varphi(\avec, \bvec)\big(\int_{\Sset}V_D(\svec')\phi(\svec', \zvec_{\zvec, \bvec}')\de \svec'- \int_{\Sset} V^*(\svec')p(\svec'|\svec, \avec)\de \svec'  \big)\beta(\zvec, \bvec)\de \zvec \de \bvec}{\int_{\Sset, \Aset}\psi(\svec, \zvec)\varphi(\avec, \bvec)\beta(\zvec, \bvec) \de \zvec \de \bvec}\bigg]  \pi(\avec | \svec) \de \avec  \nonumber  \Bigg| \nonumber  \\
		& \leq & \Bigg|\EV_D\bigg[ \int_{\Aset}\underbrace{\frac{\int_{\Sset \times \Aset}\psi(\svec, \zvec)\varphi(\avec, \bvec)\big( R_{\zvec, \bvec}  - \overline{R}(\svec,\avec)\big)\beta(\zvec, \bvec)\de \zvec \de \bvec}{\int_{\Sset, \Aset}\psi(\svec, \zvec)\varphi(\avec, \bvec)\beta(\zvec, \bvec) \de \zvec \de \bvec}}_\text{A}  \pi(\avec | \svec) \de \avec \bigg]\nonumber\Bigg| \\
		&  +  & \gamma  \Bigg|\int_{\Aset}\underbrace{\EV_D\bigg[\frac{\int_{\Sset \times \Aset}\psi(\svec, \zvec)\varphi(\avec, \bvec)\big(\int_{\Sset}V_D(\svec')\phi(\svec', \zvec_{\zvec, \bvec}')\de \svec'- \int_{\Sset} V^*(\svec')p(\svec'|\svec, \avec)\de \svec'  \big)\beta(\zvec, \bvec)\de \zvec \de \bvec}{\int_{\Sset, \Aset}\psi(\svec, \zvec)\varphi(\avec, \bvec)\beta(\zvec, \bvec) \de \zvec \de \bvec}\bigg]}_\text{B}  \pi(\avec | \svec) \de \avec  \nonumber  \Bigg| \nonumber  \\
		& \leq & \text{A}_\text{Bias} + \gamma \text{B}_\text{Bias}. \label{equation:twoterms}
	\end{eqnarray}
	It is evident that the term \textsl{A} is the Nadaraya-Watson kernel regression, as it is possible to observe in the beginnin of the proof at page twelve of \cite{tosatto_upper_2020}, therefore Theorem~\ref{theorem:biasnadaraya} applies
	\begin{eqnarray}
		\text{A}_\text{Bias} & = & \quad \frac{L_R \sum_{k=1}^d \hvec_k \Bigg(\prod_{i\neq k}^d \e{\frac{L_{\beta}^2 \hvec_i^2}{2}}\Bigg(1 +  \erf\bigg(\frac{\hvec_i L_{\beta}}{\sqrt{2}}\bigg) \Bigg)\Bigg)  \Bigg( \frac{1}{\sqrt{2 \pi} } + L_{\beta}\hvec_k \frac{\e{\frac{L_{\beta}^2 \hvec_k^2}{2}}}{2}\Bigg(1 + \erf\bigg(\frac{\hvec_k L_{\beta}}{\sqrt{2}}\bigg) \Bigg)\Bigg)}{\prod_{i=1}^d  \e{\frac{L_{\beta}^2 h_i^2}{2}}\Bigg(1 - \erf\bigg(\frac{\hvec_i L_{\beta}}{\sqrt{2}}\bigg) \Bigg)}\nonumber,  \label{bias}
	\end{eqnarray}
	where $\hvec = [\hvec_{\psi},\hvec_{\varphi}]$ and $d = d_s + d_a$.	\\
	Returning to the estimate of $\text{B}_\text{Bias}$
	\begin{eqnarray}
		& &  \Bigg|\int_{\Aset}\EV_D\bigg[\frac{\int_{\Sset \times \Aset}\psi(\svec, \zvec)\varphi(\avec, \bvec)\big(\int_{\Sset}V_D(\svec')\phi(\svec', \zvec_{\zvec, \bvec}')\de \svec'- \int_{\Sset} V^*(\svec')p(\svec'|\svec, \avec)\de \svec'  \big)\beta(\zvec, \bvec)\de \zvec \de \bvec}{\int_{\Sset, \Aset}\psi(\svec, \zvec)\varphi(\avec, \bvec)\beta(\zvec, \bvec) \de \zvec \de \bvec}\bigg] \pi(\avec | \svec) \de \avec  \nonumber  \Bigg| \nonumber \\
		&= &  \Bigg|\int_{\Aset}\frac{\int_{\Sset \times \Aset}\psi(\svec, \zvec)\varphi(\avec, \bvec)\big(\int_{\Sset}\EV\big[V_D(\svec')\phi(\svec', \zvec_{\zvec, \bvec}') \big] \de\svec'- \int_{\Sset} V^*(\svec')p(\svec'|\svec, \avec)\de \svec'  \big)\beta(\zvec, \bvec)\de \zvec \de \bvec}{\int_{\Sset, \Aset}\psi(\svec, \zvec)\varphi(\avec, \bvec)\beta(\zvec, \bvec) \de \zvec \de \bvec} \pi(\avec | \svec) \de \avec  \nonumber  \Bigg| \nonumber
	\end{eqnarray}

	One my ask whether the terms in  $\EV[V_D(\svec')\phi(\svec', \zvec_{\zvec, \bvec}')]$ are uncorrelated. The answer it is affirmative, since, even if $V_D$ depends by $\zvec_{\zvec, \bvec}$ (integral in Equation~\eqref{equation:infnpbe}), this corresponds only in the variation of a single point in the integral, and therefore the overall estimate does not change. This argument, however, does not immediately hold for the case of an infinitesimal bandwidth, and therefore we provide the results for that case separately.
	\paragraph{For Finite Bandwidth:}
	\begin{eqnarray}
		& &  \Bigg|\int_{\Aset}\frac{\int_{\Sset \times \Aset}\psi(\svec, \zvec)\varphi(\avec, \bvec)\big(\int_{\Sset}\EV\big[V_D(\svec')\phi(\svec', \zvec_{\zvec, \bvec}')\big]\de \svec'- \int_{\Sset} V^*(\svec')p(\svec'|\svec, \avec)\de \svec'  \big)\beta(\zvec, \bvec)\de \zvec \de \bvec}{\int_{\Sset, \Aset}\psi(\svec, \zvec)\varphi(\avec, \bvec)\beta(\zvec, \bvec) \de \zvec \de \bvec} \pi(\avec | \svec) \de \avec  \nonumber  \Bigg| \nonumber\\
		& \leq & \max_{\svec, \avec}\Bigg|\frac{\int_{\Sset \times \Aset}\psi(\svec, \zvec)\varphi(\avec, \bvec)\big(\int_{\Sset\times \Sset}\overline{V}(\zvec')\phi(\zvec', \svec')p(\svec'|\svec,\avec)\de \svec'\de \zvec'- \int_{\Sset} V^*(\svec')p(\svec'|\svec, \avec)\de \svec'  \big)\beta(\zvec, \bvec)\de \zvec \de \bvec}{\int_{\Sset, \Aset}\psi(\svec, \zvec)\varphi(\avec, \bvec)\beta(\zvec, \bvec) \de \zvec \de \bvec}\Bigg| \nonumber \\
		& = & \max_{\svec, \avec}\Bigg|\frac{\int_{\Sset \times \Aset}\psi(\svec, \zvec)\varphi(\avec, \bvec)\big(\int_{\Sset}\int_{\Sset}\big(\overline{V}(\zvec')\phi(\zvec', \svec')-V^*(\svec')\big)p(\svec'|\svec,\avec)\de \svec'\de \zvec'  \big)\beta(\zvec, \bvec)\de \zvec \de \bvec}{\int_{\Sset, \Aset}\psi(\svec, \zvec)\varphi(\avec, \bvec)\beta(\zvec, \bvec) \de \zvec \de \bvec}\Bigg| \nonumber \\
		& \leq & \max_{\svec, \avec, \svec'}\Bigg|\frac{\int_{\Sset \times \Aset}\psi(\svec, \zvec)\varphi(\avec, \bvec)\big(\int_{\Sset}\overline{V}(\zvec')\phi(\zvec', \svec')-V^*(\svec')\de \zvec'  \big)\beta(\zvec, \bvec)\de \zvec \de \bvec}{\int_{\Sset, \Aset}\psi(\svec, \zvec)\varphi(\avec, \bvec)\beta(\zvec, \bvec) \de \zvec \de \bvec}\Bigg| \nonumber \\
		& = & \max_{\svec, \avec, \svec'}\Bigg|\frac{\int_{\Sset \times \Aset}\psi(\svec, \zvec)\varphi(\avec, \bvec)\beta(\zvec, \bvec)\de \zvec \de \bvec}{\int_{\Sset, \Aset}\psi(\svec, \zvec)\varphi(\avec, \bvec)\beta(\zvec, \bvec) \de \zvec \de \bvec}\bigg(\int_{\Sset}\overline{V}(\zvec')\phi(\zvec', \svec')-V^*(\svec')\de \zvec'  \bigg) \Bigg|\nonumber \\
		& = & \max_{\svec, \avec, \svec'}\bigg|\int_{\Sset}\overline{V}(\zvec')\phi(\zvec', \svec')-V^*(\svec')\de \zvec'  \bigg|\nonumber \\
		& = & \max_{\svec, \avec, \svec'}\bigg|\int_{\Sset}\overline{V}(\svec'+\lvec)\phi(\svec+\lvec, \svec')-V^*(\svec')\de \lvec  \bigg|. \label{equation:finitesamples}
	\end{eqnarray}
	Note that
	\begin{equation}
		\phi(\svec' + \lvec, \svec') = \prod_{i=1}^{d_s}\frac{e^{- \frac{l_i^2}{2h^2_{\phi,i}}}}{\sqrt{2\pi h^2_{\phi,i}}}, \nonumber
	\end{equation}
	thus
	\begin{eqnarray}
		&  &  \max_{\svec, \avec, \svec'}\bigg|\int_{\Sset}\overline{V}(\svec'+\lvec)\phi(\svec+\lvec, \svec')-V^*(\svec')\de \lvec  \bigg|\nonumber\\
		& \leq & \max_{\svec, \avec, \svec'}\bigg| \overline{V}(\svec')-V^*(\svec')\bigg| + \int_{\Sset} L_{V}\bigg(\sum_{i=1}^{d_s}|l_i|\bigg)\prod_{i=1}^{d_s}\frac{e^{- \frac{l_i^2}{2h^2_{\phi,i}}}}{\sqrt{2\pi h^2_{\phi,i}}}\de. \lvec \nonumber
	\end{eqnarray}
	Using Proposition~\ref{prodsumint}
	\begin{eqnarray}
		&  & \max_{\svec, \avec, \svec'}\bigg| \overline{V}(\svec')-V^*(\svec')\bigg| + L_{V} \int_{\Sset} \bigg(\sum_{i=1}^{d_s}|l_i|\bigg)\prod_{i=1}^{d_s}\frac{e^{- \frac{l_i^2}{2h^2_{\phi,i}}}}{\sqrt{2\pi h^2_{\phi,i}}}\de \lvec \nonumber \\
		& =  & \max_{\svec, \avec, \svec'}\bigg| \overline{V}(\svec')-V^*(\svec')\bigg| +L_{V} \sum_{k=1}^{d_s}\bigg(\prod_{i\neq k}^{d_s}\int_{-\infty}^{+\infty} \frac{e^{- \frac{l_i^2}{2h^2_{\phi,i}}}}{\sqrt{2\pi h^2_{\phi,i}}}\de l_i\bigg)\int_{-\infty}^{+\infty}|l_k| \frac{e^{- \frac{l_k^2}{2h^2_{\phi,k}}}}{\sqrt{2\pi h^2_{\phi,k}}} \de l_k \nonumber \\
		& =  & \max_{\svec, \avec, \svec'}\bigg| \overline{V}(\svec')-V^*(\svec')\bigg| +L_{V}2 \sum_{k=1}^{d_s}\int_{0}^{+\infty}l_k \frac{e^{- \frac{l_k^2}{2h^2_{\phi,k}}}}{\sqrt{2\pi h^2_{\phi,k}}} \de l_k \nonumber \\
		& =  & \max_{\svec, \avec, \svec'}\bigg| \overline{V}(\svec')-V^*(\svec')\bigg| +L_{V} \sum_{k=1}^{d_s}\frac{h_{\phi,k}}{\sqrt{2 \pi}} \nonumber \\
	\end{eqnarray}
	which means that when $\hvec$ not infinitesimal
	\begin{eqnarray}
		& & \bigg| \overline{V}(\svec)-V^*(\svec)\bigg| \leq \text{A}_\text{Bias} + \gamma \bigg(\max_{\svec, \avec, \svec'}\bigg| \overline{V}(\svec')-V^*(\svec')\bigg| +L_{V} \sum_{k=1}^{d_s}\frac{h_{\phi,k}}{\sqrt{2 \pi}}\bigg).  \nonumber
	\end{eqnarray}
	It is however known that $\big|\overline{V}(\svec)-V^*(\svec)\big| \leq 2\frac{R_{\text{max}}}{1-\gamma}$, thus
	\begin{eqnarray}
		& & \bigg| \overline{V}(\svec)-V^*(\svec)\bigg| \leq \text{A}_\text{Bias} + \gamma \bigg(\max_{\svec, \avec, \svec'}\bigg| \overline{V}(\svec')-V^*(\svec')\bigg| +L_{V} \sum_{k=1}^{d_s}\frac{h_{\phi,k}}{\sqrt{2 \pi}}\bigg)  \label{equation:recursion1}  \\
		& & \bigg| \overline{V}(\svec)-V^*(\svec)\bigg| \leq \text{A}_\text{Bias} + \gamma \bigg(2\frac{R_{\text{max}}}{1-\gamma} +L_{V} \sum_{k=1}^{d_s}\frac{h_{\phi,k}}{\sqrt{2 \pi}}\bigg) \nonumber \\
		\implies & &  \bigg| \overline{V}(\svec)-V^*(\svec)\bigg| \leq \text{A}_\text{Bias} + \gamma \bigg(\text{A}_\text{Bias} + \gamma \bigg(2\frac{R_{\text{max}}}{1-\gamma} +L_{V} \sum_{k=1}^{d_s}\frac{h_{\phi,k}}{\sqrt{2 \pi}}\bigg) +L_{V} \sum_{k=1}^{d_s}\frac{h_{\phi,k}}{\sqrt{2 \pi}}\bigg)  \qquad \text{using Equation~\eqref{equation:recursion1}} \nonumber \\
		\implies & &  \bigg| \overline{V}(\svec)-V^*(\svec)\bigg| \leq \sum_{t=0}^{\infty}\gamma^t \bigg(\text{A}_\text{Bias} + \gamma L_{V} \sum_{k=1}^{d_s}\frac{h_{\phi,k}}{\sqrt{2 \pi}} \bigg)  \qquad \text{using Equation~\eqref{equation:recursion1}} \nonumber \\
		\implies & &  \bigg| \overline{V}(\svec)-V^*(\svec)\bigg| \leq  \frac{1}{1-\gamma} \bigg(\text{A}_\text{Bias} + \gamma L_{V} \sum_{k=1}^{d_s}\frac{h_{\phi,k}}{\sqrt{2 \pi}} \bigg). \nonumber 
	\end{eqnarray}
	\paragraph{For Infinitesimal Bandwidth:}
	In the case of an infinitesimal bandwidth note that, even if $V_D$ and $\phi$ are correlated the overall integral reduces only on a single point, and the same argument made in the case of finite bandwidth applies,
	\begin{eqnarray}
		\int_{\Sset}\EV\big[V_D(\svec')\phi(\svec', \zvec_{\zvec, \bvec}')\big]\de  \svec' = \EV\bigg[\int_{\Sset}V_D(\svec')\phi(\svec', \zvec_{\zvec, \bvec}'\de  \svec') \de \svec' \bigg] = \EV\big[V_D( \zvec_{\zvec, \bvec}')\big]
		=  \int_{\Sset} \overline{V}_D(\svec')p(\svec'|\svec,\avec)\de \svec'. \nonumber
	\end{eqnarray}
	It follows that, proceeding similarly to Equation~\eqref{equation:finitesamples}, we obtain
	\begin{eqnarray}
		\big|\EV_{D}[\overline{V}_D(\svec)] - V^*(\svec)\big| \leq \max_{\svec, \avec, \svec'}\bigg|\overline{V}(\svec')-V^*(\svec') \bigg|,
	\end{eqnarray}
	which yields
	\begin{equation}
		\bigg| \overline{V}(\svec)-V^*(\svec)\bigg| \leq  \frac{1}{1-\gamma} \text{A}_\text{Bias}.
	\end{equation} 
\end{proof}

\section{Empirical Evaluation Detail}

\subsection{Linear Quadratic Regulator Experiment}\label{appendix:gradient_lqr}

Here we detail the experiment presented in Figure~\ref{figure:gradient}. We use a discrete infinite-horizon discounted Linear Quadratic Regulator system of the form
\begin{align}
\max_{\vec{x}_t, \vec{u}_t} \, & J = \, \frac{1}{2} \sum_{t=0}^{\infty} \gamma^t \left( \vec{x}_t^{\top} \mathbf{Q} \vec{x}_t + \vec{u}_t^{\top} \mathbf{R} \vec{u}_t \right) \nonumber \\
%\st
& \vec{x}_{t+1}  =  \Avec \vec{x}_t + \mathbf{B} \vec{u}_t \quad \forall t, \nonumber 
\end{align}
where $\vec{x}_t \in \mathbb{R}^{d_{x}}$, $\vec{u}_t \in \mathbb{R}^{d_u}$, $\mathbf{Q} \in \mathbb{R}^{d_x \times d_x}$, $\mathbf{R} \in \mathbb{R}^{d_u \times d_u}$, $\Avec \in \mathbb{R}^{d_x \times d_x}$, $\mathbf{B} \in \mathbb{R}^{d_x \times d_u}$, $\gamma \in [0, 1)$ and $\vec{x}_0$ given. 

In this example we use consider a 2-dimensional problem with the following quantities
\begin{align*}
\Avec & =\left[\begin{array}{cc}
1.2 & 0\\
0 & 1.1
\end{array}\right] \quad \quad \quad  \mathbf{B}=\left[\begin{array}{cc}
0.1 & 0\\
0 & 0.2
\end{array}\right]\\
\mathbf{Q} & =\left[\begin{array}{cc}
-0.5 & 0\\
0 & -0.25
\end{array}\right] \quad \mathbf{R}=\left[\begin{array}{cc}
0.01 & 0\\
0 & 0.01
\end{array}\right]\\
\vec{x}_0 & = \left[ \begin{array}{c}
1 \\
1
\end{array} \right]
\quad \quad 
\gamma  = 0.9.
\end{align*}

For this LQR problem we impose a linear controller as a diagonal matrix
\begin{align}
\mathbf{K} = \left[ \begin{array}{cc}
k_1 & 0  \\
0 & k_2
\end{array} \right]. \label{eq:lqr_diagonal_policy}
\end{align}

\subsubsection{Deterministic Experiment}

For each dataset we run 100 trajectories of 30 steps. Each trajectory is generated by following the dynamics of the described LQR and using at each time step a fixed policy initialized as
\begin{align}
\mathbf{K} = \left[ \begin{array}{cc}
k_1 + \varepsilon & \varepsilon  \\
\varepsilon & k_2 + \varepsilon 
\end{array} \right], \; \varepsilon \sim \mathcal{N}(0, 1), \nonumber
\end{align}
where $k_1 = 0.7$ and $k_2 = -0.7$.

NOPG-D optimized for each dataset a policy encoded as in \eqref{eq:lqr_diagonal_policy} with: learning rate $0.5$ with ADAM optimizer; bandwidths (on average) for the state space  $\vec{h}_{\psi} = [0.03, 0.05]$ and for the action space $\vec{h}_{\varphi} = [0.33, 0.27]$; discount factor $\gamma=0.9$; and keeping 5 elements per row after sparsification of the $\mathbf{P}$ matrix.

DPG optimized for each dataset a policy encoded as in \eqref{eq:lqr_diagonal_policy} with: learning rate $0.5$ with ADAM optimizer; $Q$-function encoded as $Q(\vec{x}, \vec{u}) = \vec{x}^{\top} \mathbf{Q} \vec{x} + \vec{u}^{\top} \mathbf{R} \vec{u}$ (with $\mathbf{Q}$ and $\mathbf{R}$ to be learned); discount factor $\gamma=0.9$; two target networks are kept to stabilize learning and soft-updated using $\tau = 0.01$ (similar to DDPG).

\subsubsection{Stochastic Experiment} 

For each dataset we run 100 trajectories of 30 steps. Each trajectory is generated by following the dynamics of the described LQR, and using at each time step a stochastic policy as
\begin{align}
\vec{u}_t = \mathbf{K}  \vec{x}_t + \vec{\varepsilon}, \; \vec{\varepsilon} \sim \mathcal{N} \left( \vec{\mu} = \vec{0}, \mathbf{\Sigma} = \textrm{diag}( 0.01, 0.01) \right), \label{eq:lqr_diagonal_policy_stochastic}
\end{align}
where $\mathbf{K} = \textrm{diag}(0.35, -0.35)$.

NOPG-S optimized for each dataset a policy encoded as in \eqref{eq:lqr_diagonal_policy_stochastic} with: learning rate $0.25$ with ADAM optimizer; bandwidths (on average) for the state space $\vec{h}_{\psi} = [0.008, 0.003]$ and for the action space $\vec{h}_{\varphi} = [0.02, 0.02]$; discount factor $\gamma=0.9$; and keeping 10 elements per row after sparsification of the $\mathbf{P}$ matrix.

PWIS optimized for each dataset a policy encoded as in \eqref{eq:lqr_diagonal_policy_stochastic} with: learning rate $2.5 \times 10^{-4}$ with ADAM optimizer; and discount factor $\gamma=0.9$.

\subsection{Other Experiments Configurations}\label{appendix:experiments_configs}

We use a policy encoded as neural network with parameters $\vec{\theta}$. A deterministic policy is encoded with a neural network $\action = f_{\vec{\theta}}(\state)$. The stochastic policy is encoded as a Gaussian distribution with parameters determined by a neural network with two outputs, the mean and covariance. In this case we represent by $f_{\vec{\theta}}(\state)$ the slice of the output corresponding to the mean and by $g_{\vec{\theta}}(\state)$ the part of the output corresponding to the covariance. 
%In some algorithms we use a latent representation as part of the policy and value function networks, denoted by $\vec{z} = h_{\vec{\theta}}(\state)$, which is then linearly transformed to give the Gaussian policy parameters or to compute the value function.

NOPG can be described with the following hyper-parameters

\begin{longtable}{l l}
	\textbf{NOPG Parameters} & Meaning\\
	\hline
	dataset sizes & number of samples contained in the dataset used for training \\        
	discount factor $\gamma$ & usual discount factor in infinite horizon MDP \\
	state $\vec{h}_{\textrm{factor}}$ & constant used to decide the bandwidths for the state-space \\
	action $ \vec{h}_{\textrm{factor}}$ & constant used to decide the bandwidths for the action-space \\
	policy & parametrization of the policy\\
	policy output & how is the output of the policy encoded \\               
	learning rate  & the learning rate and the gradient ascent algorithm used \\
	$N_{\pi}^{\textrm{MC}}$ (NOPG-S) & number of samples drawn to compute the integral $\bm{\varepsilon}_{\pi}(\state)$ with MonteCarlo sampling \\
	$N_{\phi}^{\textrm{MC}}$ & number of samples drawn to compute the integral over the next state $\int \phi(\state')\de \state'$\\
	$N_{\mu_0}^{\textrm{MC}}$ & number of samples drawn to compute the integral over the initial \\
	&  distribution $\int \hat{V}_{\pi}(\state) \mu_{0}(\state) \de \state$ \\
	policy updates & number of policy updates before returning the optimized policy \\
	\hline
\end{longtable}

A  few considerations about NOPG parameters. If $N_{\phi}^{\textrm{MC}}=1$ we use the mean of the kernel $\phi$ as a sample to approximate the integral over the next state. When optimizing a stochastic policy represented by a Gaussian distribution, we set and linearly decay the variance over the policy optimization procedure. The kernel bandwidths are computed in two steps: first we find the best bandwidth for each dimension of the state and action spaces using cross validation; second we multiply each bandwidth by an empirical constant factor ($\vec{h}_{\textrm{factor}}$). This second step is important to guarantee that the state and action spaces do not have a zero density. For instance, in a continuous action environment, when sampling actions from a uniform grid we have to guarantee that the space between the grid points have some density. The problem of estimating the bandwidth in kernel density estimation is well studied, but needs to be adapted to the problem at hand, specially with a low number of samples. We found this approach to work well for our experiments but it still can be improved.

\subsubsection{Pendulum with Uniform Dataset}
Tables~\ref{tab:pendulum_uniform} and~\ref{tab:pendulum_uniform_configs} describe the hyper-parameters used to run the experiment shown in the first plot of Figure~\ref{figure:comparison}.

\paragraph{Dataset Generation}
The dataset have been generated using a grid over the state-action space $\theta, \dot{\theta}, u$, where $\theta$ and $\dot{\theta}$ are respectively angle and angular velocity of the pendulum, and $u$ is the torque applied. 
In Table~\ref{tab:pendulum_uniform} are enumerated the different dataset used.

\begin{table}[H]
\begin{center}
    \begin{tabular}{c c c c}
        $\#\theta$ & $\#\dot{\theta}$ & $\#u$ & Sample size \\ \hline
        $10$ & $10$ & $2$ & $200$ \\
        $15$ & $15$ & $2$ & $450$ \\
        $20$ & $20$ & $2$ & $800$ \\
        $25$ & $25$ & $2$ & $1250$ \\
        $30$ & $30$ & $2$ & $1800$ \\
        $40$ & $40$ & $2$ & $3200$ \\
        \hline
    \end{tabular}
    \end{center}

    \caption[Pendulum uniform grid dataset configurations]{\textbf{Pendulum uniform grid dataset configurations} This table shows the level of discretization for each dimension of the state space ($\#\theta$ and $\#\dot{\theta}$) and the action space ($\#u$). Each line corresponds to a uniformly sampled dataset, where $\theta \in [-\pi, \pi]$, $\dot{\theta} \in [-8, 8]$ and $u \in [-2, 2]$. The entries under the states' dimensions and action dimension correspond to how many linearly spaced states or actions are to be queried from the corresponding intervals. The Cartesian product of states and actions dimensions is taken in order to generate the state-action pairs to query the environment transitions. The rightmost column indicates the total number of corresponding samples.}
    \label{tab:pendulum_uniform}
\end{table}

\paragraph{Algorithm details.}

The configuration used for NOPG-D and NOPG-S are listed in Table~\ref{tab:pendulum_uniform_configs}.
\begin{table}[H]
\begin{center}

    \begin{tabular}{l l}
        \textbf{NOPG} & \\
        \hline
        discount factor $\gamma$ & 0.97 \\
        state $\vec{h}_{\textrm{factor}}$ & $1.0$ $1.0$  $1.0$   \\
        action $ \vec{h}_{\textrm{factor}}$ & $50.0$ \\
        policy & neural network parameterized by $\vec{\theta}$\\
               & 1 hidden layer, 50 units, ReLU activations \\
        policy output & $2 \tanh(f_{\vec{\theta}}(\state))$ (NOGP-D) \\
                      & $\mu = 2 \tanh(f_{\vec{\theta}}(\state))$, $\sigma = \textrm{sigmoid}(g_{\vec{\theta}}(\state))$ (NOGP-S) \\               
        learning rate  & $10^{-2}$ with ADAM optimizer \\
        $N_{\pi}^{\textrm{MC}}$ (NOPG-S) & 15 \\
        $N_{\phi}^{\textrm{MC}}$ & 1 \\
        $N_{\mu_0}^{\textrm{MC}}$ & (non applicable) fixed initial state \\           
        policy updates & $1.5 \cdot 10^3$ \\
        \hline
    \end{tabular}
    \end{center}

    \caption[NOPG configurations for the Pendulum uniform grid experiment]{\textbf{NOPG configurations for the Pendulum uniform grid experiment}
%    This table contains the configurations of NOPG-D and NOPG-S needed to replicate the results from the Pendulum uniform grid experiment (\ref{tab:pendulum_uniform_results}).
    }
    \label{tab:pendulum_uniform_configs}
\end{table}

\subsubsection{Pendulum with Random Agent}
The following table shows the hyper-parameters used for generating the second plot starting from the left in Figure~\ref{figure:comparison}
\begin{center}
    \begin{longtable}{l l}
        \textbf{NOPG} & \\
        \hline
        dataset sizes & $10^2$, $5 \cdot 10^2$, $10^3$, $1.5 \cdot 10^3$, $2 \cdot 10^3$, $3 \cdot 10^3$, \\
                      & $5 \cdot 10^3$, $7 \cdot 10^3$, $9 \cdot 10^3$, $10^4$ \\
        discount factor $\gamma$ & $0.97$ \\
        state $\vec{h}_{\textrm{factor}}$ & $1.0$ $1.0$ $1.0$  \\
        action $ \vec{h}_{\textrm{factor}}$ & $25.0$ \\
        policy & neural network parameterized by $\vec{\theta}$\\
               & $1$ hidden layer, $50$ units, ReLU activations \\
        policy output & $2 \tanh(f_{\vec{\theta}}(\state))$ (NOGP-D) \\
                      & $\mu = 2 \tanh(f_{\vec{\theta}}(\state))$, $\sigma = \textrm{sigmoid}(g_{\vec{\theta}}(\state))$ (NOGP-S) \\               
        learning rate  & $10^{-2}$ with ADAM optimizer \\
        $N_{\pi_0}^{\textrm{MC}}$ (NOPG-S) & $10$ \\
        $N_{\phi}^{\textrm{MC}}$ & $1$ \\
        $N_{\mu_0}^{\textrm{MC}}$ & (non applicable) fixed initial state \\        
        policy updates & $2 \cdot 10^3$ \\
        \hline
        & \\ & \\
        
        \textbf{DDPG} & \\
        \hline
        discount factor $\gamma$ & $0.97$ \\
        rollout steps & $1000$ \\
        actor  & neural network parameterized by $\vec{\theta}_{\textrm{actor}}$\\
               & $1$ hidden layer, $50$ units, ReLU activations \\
        actor output & $2 \tanh(f_{\vec{\theta}_{\textrm{actor}}}(\state))$ \\
        actor learning rate & $10^{-3}$ with ADAM optimizer \\
        critic  & neural network parameterized by $\vec{\theta}_{\textrm{critic}}$\\
                & $1$ hidden layer, $50$ units, ReLU activations \\
        critic output & $f_{\vec{\theta}_{\textrm{critic}}}(\state, \action)$ \\
        critic learning rate & $10^{-2}$ with ADAM optimizer \\        
        soft update & $\tau = 10^{-3}$ \\
        policy updates & $3 \cdot 10^5$ \\
        \hline        
        & \\ & \\
        
        \textbf{DDPG Offline} & \\
        \hline
        dataset sizes & $10^2$, $5 \cdot 10^2$, $10^3$, $2 \cdot 10^3$, $5 \cdot 10^3$, $7.5 \cdot 10^3$, \\
                      &  $10^4$, $1.2 \cdot 10^4$, $1.5 \cdot 10^4$, $2 \cdot 10^4$, $2.5 \cdot 10^4$ \\
        discount factor $\gamma$ & $0.97$ \\
        actor  & neural network parameterized by $\vec{\theta}_{\textrm{actor}}$\\
               & $1$ hidden layer, $50$ units, ReLU activations \\
        actor output & $2 \tanh(f_{\vec{\theta}_{\textrm{actor}}}(\state))$ \\
        actor learning rate & $10^{-2}$ with ADAM optimizer \\
        critic  & neural network parameterized by $\vec{\theta}_{\textrm{critic}}$\\
                & $1$ hidden layer, $50$ units, ReLU activations \\
        critic output & $f_{\vec{\theta}_{\textrm{critic}}}(\state, \action)$ \\
        critic learning rate & $10^{-2}$ with ADAM optimizer \\
        soft update & $\tau = 10^{-3}$ \\
        policy updates & $2 \cdot 10^3$ \\
        \hline
        & \\ & \\

        \textbf{PWIS} & \\
        \hline
        dataset sizes & $10^2$, $5 \cdot 10^2$, $10^3$, $2 \cdot 10^3$, $5 \cdot 10^3$, $7.5 \cdot 10^3$, \\
                      &  $10^4$, $1.2 \cdot 10^4$, $1.5 \cdot 10^4$, $2 \cdot 10^4$, $2.5 \cdot 10^4$ \\        
        discount factor $\gamma$ & $0.97$ \\
        policy & neural network parameterized by $\vec{\theta}$\\
               & $1$ hidden layer, $50$ units, ReLU activations \\
        policy output & $\mu = 2 \tanh(f_{\vec{\theta}}(\state))$, $\sigma = \textrm{sigmoid}(g_{\vec{\theta}}(\state))$ \\               
        learning rate  & $10^{-2}$ with ADAM optimizer \\
        policy updates & $2 \cdot 10^3$ \\
        \hline

        \caption[Algorithms configurations for the Pendulum random data experiment]{\textbf{Algorithms configurations for the Pendulum random data experiment}}
        \label{tab:pendulum_random_configs}
            
    \end{longtable}
\end{center}

\subsubsection{Cart-pole with Random Agent}
The following tables show the hyper-parameters used to generate the third plot in Figure~\ref{figure:comparison}.
\begin{center}
    \begin{longtable}{l l}
        \textbf{NOPG} & \\
        \hline
        dataset sizes & $10^2$, $2.5 \cdot 10^2$, $5 \cdot 10^2$, $10^3$, $1.5 \cdot 10^3$, $2.5 \cdot 10^3$,  \\
                      & $3 \cdot 10^3$, $5 \cdot 10^3$, $6 \cdot 10^3$, $8 \cdot 10^3$, $10^4$ \\        
        discount factor $\gamma$ & $0.99$ \\
        state $\vec{h}_{\textrm{factor}}$ & $1.0$ $1.0$ $1.0$  \\
        action $ \vec{h}_{\textrm{factor}}$ & $20.0$ \\
        policy & neural network parameterized by $\vec{\theta}$\\
               & $1$ hidden layer, $50$ units, ReLU activations \\
        policy output & $5 \tanh(f_{\vec{\theta}}(\state))$ (NOGP-D) \\
                      & $\mu = 5 \tanh(f_{\vec{\theta}}(\state))$, $\sigma = \textrm{sigmoid}(g_{\vec{\theta}}(\state))$ (NOGP-S) \\               
        learning rate  & $ \cdot 10^{-2}$ with ADAM optimizer \\
        $N_{\pi}^{\textrm{MC}}$ (NOPG-S) & $10$ \\
        $N_{\phi}^{\textrm{MC}}$ & $1$ \\
        $N_{\mu_0}^{\textrm{MC}}$ & $15$ \\
        policy updates & $2 \cdot 10^3$ \\
        \hline
        & \\ & \\
        
        \textbf{DDPG} & \\
        \hline
        discount factor $\gamma$ & $0.99$ \\
        rollout steps & $1000$ \\
        actor  & neural network parameterized by $\vec{\theta}_{\textrm{actor}}$\\
               & $1$ hidden layer, $50$ units, ReLU activations \\
        actor output & $5 \tanh(f_{\vec{\theta}_{\textrm{actor}}}(\state))$ \\
        actor learning rate & $10^{-3}$ with ADAM optimizer \\
        critic  & neural network parameterized by $\vec{\theta}_{\textrm{critic}}$\\
                & $1$ hidden layer, $50$ units, ReLU activations \\
        critic output & $f_{\vec{\theta}_{\textrm{critic}}}(\state, \action)$ \\
        critic learning rate & $10^{-2}$ with ADAM optimizer \\        
        soft update & $\tau = 10^{-3}$ \\
        policy updates & $2 \cdot 10^5$ \\
        \hline        
        & \\ & \\ & \\
        
        \textbf{DDPG Offline} & \\
        \hline
        dataset sizes & $10^2$, $5 \cdot 10^2$, $10^3$, $2 \cdot 10^3$, $3.5 \cdot 10^3$, $5 \cdot 10^3$, \\
                      & $8 \cdot 10^3$,  $10^4$, $1.5 \cdot 10^4$, $2 \cdot 10^4$, $2.5 \cdot 10^4$ \\        
        discount factor $\gamma$ & $0.99$ \\
        actor  & neural network parameterized by $\vec{\theta}_{\textrm{actor}}$\\
               & $1$ hidden layer, $50$ units, ReLU activations \\
        actor output & $5 \tanh(f_{\vec{\theta}_{\textrm{actor}}}(\state))$ \\
        actor learning rate & $10^{-2}$ with ADAM optimizer \\
        critic  & neural network parameterized by $\vec{\theta}_{\textrm{critic}}$\\
                & $1$ hidden layer, $50$ units, ReLU activations \\
        critic output & $f_{\vec{\theta}_{\textrm{critic}}}(\state, \action)$ \\
        critic learning rate & $10^{-2}$ with ADAM optimizer \\
        soft update & $\tau = 10^{-3}$ \\
        policy updates & $2 \cdot 10^3$ \\
        \hline
        & \\ & \\

        \textbf{PWIS} & \\
        \hline
        dataset sizes & $10^2$, $5 \cdot 10^2$, $10^3$, $2 \cdot 10^3$, $3.5 \cdot 10^3$, $5 \cdot 10^3$, \\
                      & $8 \cdot 10^3$,  $10^4$, $1.5 \cdot 10^4$, $2 \cdot 10^4$, $2.5 \cdot 10^4$ \\          
        discount factor $\gamma$ & $0.99$ \\
        policy & neural network parameterized by $\vec{\theta}$\\
               & $1$ hidden layer, $50$ units, ReLU activations \\
        policy output & $\mu = 5 \tanh(f_{\vec{\theta}}(\state))$, $\sigma = \textrm{sigmoid}(g_{\vec{\theta}}(\state))$ \\               
        learning rate  & $10^{-3}$ with ADAM optimizer \\
        policy updates & $2 \cdot 10^3$ \\
        \hline

        \caption[Algorithms configurations for the CartPole random data experiment]{\textbf{Algorithms configurations for the CartPole random data experiment}.}
        \label{tab:cartpole_random_configs}
            
    \end{longtable}
\end{center}

\subsubsection{Mountain Car with Human Demonstrator}
Here the detail of the experiment shown in Figure~\ref{figure:mountain}.
The dataset in this experiment ($10$ trajectories) has been generated by a human demonstrator. The dataset used is available in the source code provided.

\begin{center}

    \begin{longtable}{l l}
        \textbf{NOPG} & \\
        \hline
        discount factor $\gamma$ & $0.99$ \\
        state $\vec{h}_{\textrm{factor}}$ & $1.0$ $1.0$ \\
        action $ \vec{h}_{\textrm{factor}}$ & $50.0$ \\
        policy & neural network parameterized by $\vec{\theta}$\\
               & $1$ hidden layer, $50$ units, ReLU activations \\
        policy output & $1 \tanh(f_{\vec{\theta}}(\state))$ (NOGP-D) \\
                      & $\mu = 1 \tanh(f_{\vec{\theta}}(\state))$, $\sigma = \textrm{sigmoid}(g_{\vec{\theta}}(\state))$ (NOGP-S) \\               
        learning rate  & $10^{-2}$ with ADAM optimizer \\
        $N_{\pi}^{\textrm{MC}}$ (NOPG-S) & $15$ \\
        $N_{\phi}^{\textrm{MC}}$ & $1$ \\
        $N_{\mu_0}^{\textrm{MC}}$ & $15$ \\
        policy updates & $1.5 \cdot 10^3$ \\
        \hline
        \caption[NOPG configurations for the MountainCar experiment]{\textbf{NOPG configurations for the MountainCar experiment.}}
        \label{tab:mountaincar_configs}
    \end{longtable}
    
\end{center}

%\end{adjustwidth}

\end{document}